\documentclass[10pt,twocolumn,letterpaper]{article}

\usepackage{iccv}
\usepackage{times}
\usepackage{epsfig}
\usepackage{graphicx}
\usepackage{amsmath}
\usepackage{amssymb}
\usepackage{gensymb}

\usepackage[utf8]{inputenc} 
\usepackage[T1]{fontenc}    
\usepackage{url}            
\usepackage{booktabs,arydshln}       
\usepackage{amsfonts}       
\usepackage{nicefrac}       
\usepackage{microtype}      
\usepackage{adjustbox}
\usepackage{xcolor}
\usepackage{xspace}
\usepackage{multirow}
\usepackage{bbold}
\usepackage{subcaption}
\usepackage{enumitem}
\usepackage{color, colortbl}
\usepackage{soul}
\usepackage{flushend}

\makeatletter
\def\adl@drawiv#1#2#3{%
        \hskip.5\tabcolsep
        \xleaders#3{#2.5\@tempdimb #1{1}#2.5\@tempdimb}%
                #2\z@ plus1fil minus1fil\relax
        \hskip.5\tabcolsep}
\newcommand{\cdashlinelr}[1]{%
  \noalign{\vskip\aboverulesep
           \global\let\@dashdrawstore\adl@draw
           \global\let\adl@draw\adl@drawiv}
  \cdashline{#1}
  \noalign{\global\let\adl@draw\@dashdrawstore
           \vskip\belowrulesep}}
\makeatother

\definecolor{Gray}{gray}{0.9}

\definecolor{mygreen}{RGB}{169, 209, 142}

\usepackage[pagebackref=true,breaklinks=true,letterpaper=true,colorlinks,bookmarks=false]{hyperref}

\iccvfinalcopy 

\def\iccvPaperID{5408} 

\ificcvfinal\fi

\definecolor{darkgreen}{rgb}{0,0.45,0}

\newcommand\minisection[1]{\vspace{1mm}\noindent \textbf{#1}}
\newcommand{\model}{{Faster R-CNN}\xspace} 
 
\newcommand{\task}{{CycConf}\xspace} 

\begin{document}

\title{Robust Object Detection via Instance-Level Temporal Cycle Confusion}

\author{
    Xin Wang$^1$~~~~
    Thomas E. Huang$^{2}$\thanks{Equal contribution. The authors are listed in alphabetical order.}~~~~
    Benlin Liu$^{3*}$~~~~\\
    Fisher Yu$^2$~~~~
    Xiaolong Wang$^4$~~~~
    Joseph E. Gonzalez$^5$~~~~
    Trevor Darrell$^5$~~~~
    \\\\
    $^1$\normalsize Microsoft Research~~~~$^2$\normalsize ETH Z{\"u}rich~~~~
    $^3$\normalsize University of Washington~~~~$^4$\normalsize UC San Diego~~~~$^5$\normalsize UC Berkeley\\
}

\maketitle

\begin{abstract}
Building reliable object detectors that are
robust to domain shifts, such as various changes
in context, viewpoint, and object appearances,
is critical for real-world applications. In this
work, we study the effectiveness of auxiliary 
self-supervised tasks to improve the out-of-distribution generalization of object detectors. Inspired by the principle of maximum entropy, we introduce a novel self-supervised task, instance-level temporal cycle confusion (\task), which
operates on the region features of the object detectors. For each object, the task is to find the most different object proposals in the adjacent frame in a video and then cycle back to itself for self-supervision. \task encourages the object detector to explore invariant structures across instances under various motions, which leads to improved model robustness in unseen domains at test time. We observe consistent out-of-domain performance improvements when training object detectors in tandem with self-supervised tasks on various domain adaptation benchmarks with static images (Cityscapes, Foggy Cityscapes, Sim10K) and large-scale video datasets (BDD100K and Waymo open data)\footnote{The models are released at \url{https://xinw.ai/cyc-conf}}. \vspace{-3mm}
\end{abstract}

\section{Introduction}
Object detection has achieved remarkable performance on 
in-domain data~\cite{carion2020end,ren2015faster}. However, contemporary visual
perception models still suffer significant performance
degradation under domain shifts, raising concerns 
for safety critical applications such as autonomous
driving~\cite{dikmen2016autonomous,koopman2017autonomous}. 

Prior works~\cite{cai2019exploring,chen2018domain,hsu2020every,kim2019diversify,saito2019strong,zhu2019adapting} have designed domain adaptive object detectors, which align the unlabeled target domain data and labeled source domain data in the feature space to tackle the distribution shift problem.
These approaches perform well when they have access to excessive unlabeled data in the target domain.
More recently, another line of work improves model robustness to domain shift~\cite{koh2020wilds,bdd100k}, image corruption and distortion~\cite{hendrycks2018benchmarking} through pre-training~\cite{hendrycks2019using}, self-supervision~\cite{hendrycks2019self}, and data augmentation~\cite{hendrycks2019augmix}. These studies are largely conducted on image classification, and the effectiveness is unknown for structural prediction tasks like object detection.

\begin{figure}[tp]
  \centering
  \includegraphics[width=.9\linewidth]{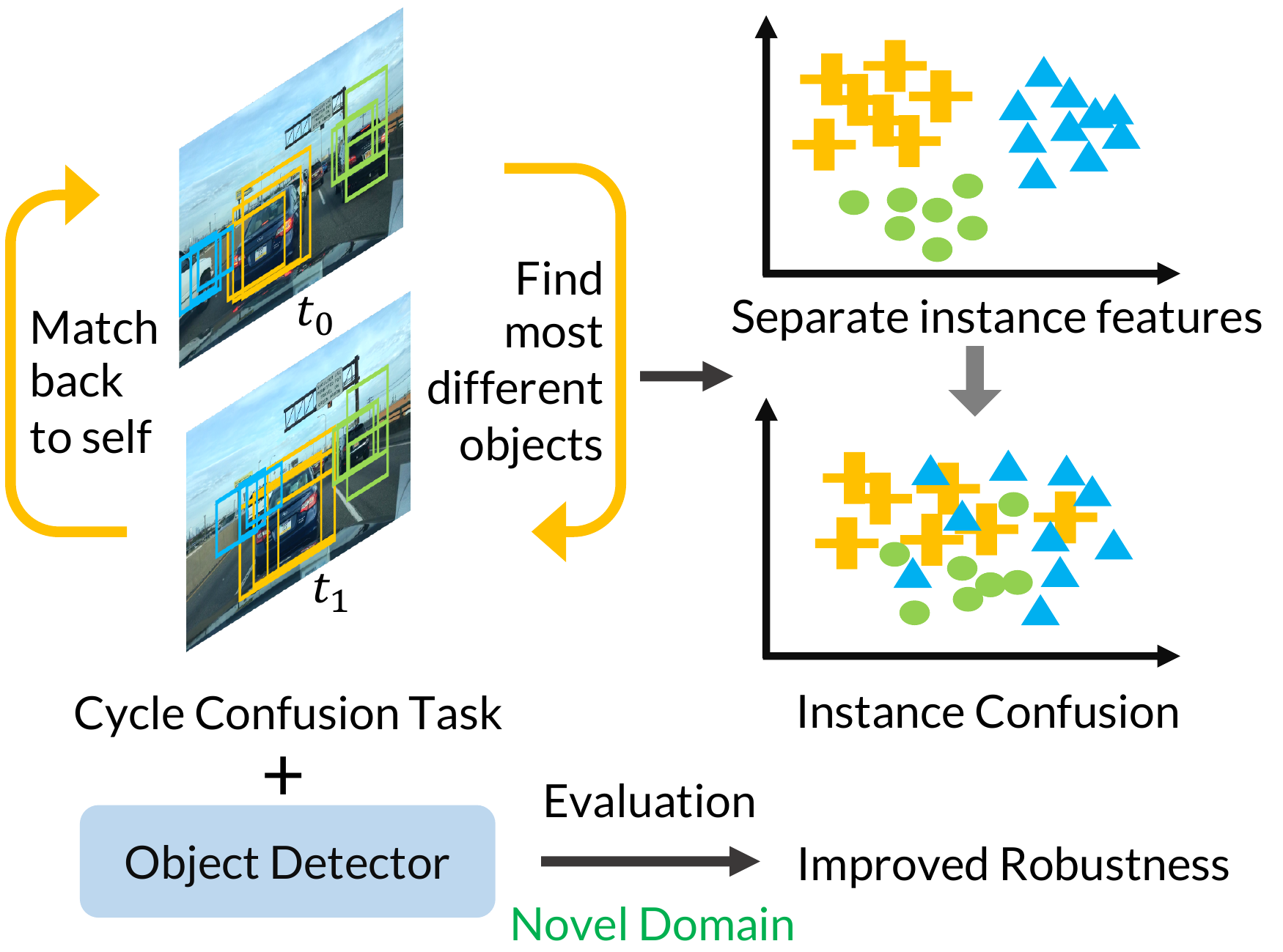}
  \caption{\small We introduce a self-supervised task, instance-level temporal cycle confusion (\task), which mixes up instance features to encourage learning invariant structures across instances.  The object detector trained in tandem with the auxiliary self-supervised task is more robust to domain shifts. \vspace{-6mm}
  }
  \label{fig:teaser}
\end{figure}

In this work, we revisit the idea of auxiliary self-supervised tasks
to improve out-of-domain generalization of object detectors. Through empirical studies, we find 
the widely used self-supervised tasks such as image rotation~\cite{gidaris2018unsupervised} and Jigsaw~\cite{noroozi2016unsupervised}, used in tandem with the fully supervised object detector in the source domain, can consistently improve the object detector's out-of-domain performance (e.g., evaluating on different datasets or different scenes) without using target data. 
Surprisingly, we also find that jointly training the object detector and the rotation task outperforms the complicated feature alignment approaches by a large margin on a range of unsupervised domain adaptation (UDA) benchmarks,
where the test domain is known during training.
These findings indicate the usage of auxiliary self-supervised tasks can be a general solution to improve model robustness under various assumptions about the amount of unlabeled target data available. 

While the findings are inspiring, we take a step further to ask, what would be a good self-supervised task for out of domain object detection? Here we introduce a new self-supervised task, instance-level cycle confusion (\task), which operates on the region features of the
object detectors as shown in Figure~\ref{fig:teaser}.
For a pair of frames in a video sequence, the \task task is to find the \emph{most different} objects through time in the frames.
Inspired by the principle of maximum entropy~\cite{jaynes1957information,jaynes1957information2}, \task mixes up the instance features 
to encourage the model to explore invariant structures across instances which may be under various motion, viewpoint and lightning conditions. In contrast to object tracking, which finds identical objects through time, \task encourages cross-instance matching which increases confusion among instances and encourages the object detector to explore the latent structures of the instances that are invariant to changing environments. Therefore, the object detector trained in tandem with \task has improved robustness to domain shifts. 

To evaluate the effectiveness of our new self-supervised task, we construct a benchmark of out-of-domain generalization for object detection using the video datasets BDD100K~\cite{bdd100k} and Waymo Open data~\cite{sun2019scalability}, which are the largest contemporary driving video datasets in the open source community. The datasets contain various object scales and diverse scenes, which is a good testbed for new model designs. In this benchmark, we consider various out-of-domain scenarios (e.g., generalization across different time of day, camera views and datasets). 
The proposed
\task task improves the baseline model by 2 to 5 points in average precision (AP50), outperforming other self-supervised tasks when evaluating the object detector in unseen domains. 
Our contributions can be summarized as follows. 
\setlist{nolistsep}
\begin{itemize}[noitemsep,leftmargin=*]
\item We show the adoption of auxiliary self-supervised tasks is a general solution to improve the robustness of object detectors to domain shifts under various assumptions. 
\item We introduce instance-level cycle confusion (\task), a novel self-supervised task on instance features, which improves the object detectors' robustness.
\item The proposed approach achieves state-of-art performances on a range of domain adaptation benchmarks. We additionally introduce an out-of-domain generalization benchmark for object detection using large-scale driving videos.
\end{itemize}

\section{Related Work}
Our work is in line with the model robustness and domain adaptation literature with an emphasis on the object detection task. The design of \task is conceptually related to the principle of maximum entropy and connects to other self-supervised tasks in the literature. 

\minisection{Domain adaptation and robustness.} Generalization
under domain shifts is a core problem in machine learning and
computer vision. Many~\cite{cai2019exploring,chen2018domain,hsu2020every, kim2019diversify,saito2019strong,su2020adapting,xu2020exploring,xu2020cross,zhao2020collaborative,zhu2019adapting} have designed domain adaptive object and used the unlabeled target 
domain data through feature alignment. Sun~\etal~\cite{sun2019unsupervised} use an auxiliary rotation task to leverage the unlabeled target domain data for semantic segmentation.
However, accessing massive target data is not feasible in many real-world applications~\cite{dikmen2016autonomous,koopman2017autonomous}. In another line, some works~\cite{geirhos2018imagenet, hendrycks2018benchmarking,recht2018cifar, shankar2019systematic} focus on testing the robustness of image classification models on out-of-domain data at inference, seeking to improve the model robustness to domain shfits~\cite{koh2020wilds,bdd100k}, image corruption and distortion~\cite{hendrycks2018benchmarking} through pre-training~\cite{hendrycks2019using}, data augmentation~\cite{hendrycks2019augmix} and self-supervision~\cite{hendrycks2019self}. 

Both Hendrycks~\etal~\cite{hendrycks2019self} and Sun~\etal~\cite{sun2019unsupervised} show that the adoption of an auxiliary rotation task improves model robustness to image corruption and domain shifts. However, they do not touch on the structural prediction tasks like object detection and do not discuss the choices of self-supervised tasks under various assumptions (e.g., the availability of test domain data).
In this work, we bridge the gap between the unsupervised domain adaptation and model robustness literature and show that using auxiliary self-supervised tasks is a general solution to improve the out-of-domain performance in different situations. We emphasize on the object detection task and design a new instance-level self-supervised task that better suits object detectors.

\minisection{Maximum entropy.} The principle of maximum entropy was founded by E. T. Jaynes in 1950s~\cite{jaynes1957information,jaynes1957information2} for statistical mechanics and information theory, which indicates that the probability distribution with the highest entropy is the one that best represents the current state of knowledge in the context of precisely stated prior data. Maximum entropy reinforcement learning (MaxEnt RL)~\cite{rawlik2012stochastic,todorov2006linearly,ziebart2010modeling} has a set of robust reinforcement learning algorithms built upon such concept. Eysenbach and Levin~\cite{eysenbach2021maximum} recently provide a theoretical justification of the robustness of MaxEnt RL under various environments. 

The design of \task is conceptually related to the principle of maximum entropy, which increases the entropy of the matching prediction probability distribution by matching the most different objects across frames. From the adversary perspective, \task makes it challenging to distinguish the instance identities and encourages the object detector to explore the invariant structure among instances, which may be under various motion, viewpoint or lightning conditions. 

\begin{figure*}[tp]
  \centering
  \includegraphics[width=\linewidth]{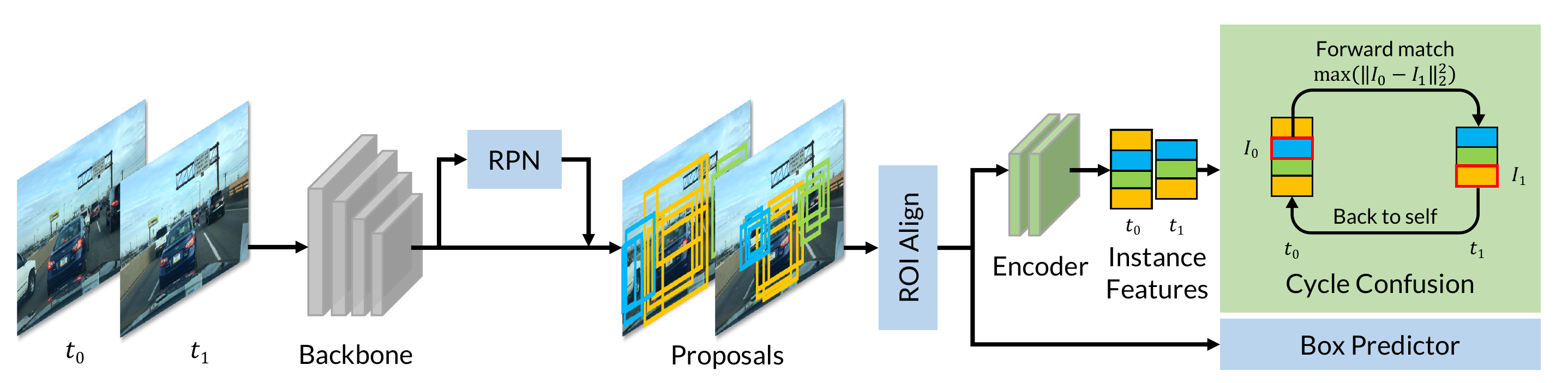}
  \caption{\small Overall model architecture. A two-stage object detector is trained in tandem with the instance-level cycle confusion (\task) task. For a pair of images, we transform the region features produced by the region proposal network (RPN) with two convolutional layers (encoder) and then use the transformed instance features as input to the cycle confusion task shown in \textbf{\textcolor{mygreen}{green}}. In the forward matching from $t_0$ to $t_1$, the matching object $I_1$ is defined as the object with maximum distance to the object $I_0$ in $t_0$ (a.k.a. the most different object). The matching object in $t_1$ cycles back to the original object in $t_0$ for self supervision. \vspace{-2mm}}
  \label{fig:arch}
\end{figure*}

\minisection{Self-supervised learning.} In self-supervised learning, a pretext task is usually designed to provide auxiliary signals to train neural networks~\cite{doersch2015unsupervised,gidaris2018unsupervised, noroozi2016unsupervised, pathak2016context,vincent2008extracting, zhang2017split}. For example, the image rotation task~\cite{gidaris2018unsupervised} applies a rotation transformation to the input image and requires the network to estimate what is the applied transformation. Besides creating pretext tasks in images, researchers have extended self-supervised learning with video data~\cite{dwibedi2019temporal,owens2016ambient,sermanet2018time,Wang_UnsupICCV2015,wang2019learning}. Dwibedi~\etal\cite{dwibedi2019temporal} adopt temporal cycle consistency to provide supervision for learning frame-level correspondence across multiple videos.  
Wang~\etal\cite{wang2019learning} introduce a self-supervised method for learning visual 
correspondence through cycle-consistency of time. 
The core idea in these approaches is to enable matching frames using
nearest neighbors in the learned per-frame embedding space, which is useful for video
alignment/correspondence tasks. To the best of our knowledge, \task is the first self-supervised task that operates on the instance features and improves the robustness of object detectors.

\section{Object Detection with Cycle Confusion}
Domain shifts, such as various weather conditions, time of day, viewpoints and geo-locations, are one of the key issues in perception systems. We design robust object detectors
that are resistant to domain shifts. Other aspects of
model robustness, such as adversarial examples and corruptions, are not covered in the scope of this work.

We improve the out-of-domain generalization of object detectors, which mitigates the performance degeneration when evaluating the
object detectors in unseen test domains different from the source domain. That is, an object detector $\mathcal{F}$ is trained on a
source domain $\mathcal{D}_s$ (e.g., daytime,  highway) and is evaluated on a different test domain $\mathcal{D}_t$ (e.g., night time, city street), where the test domain is \emph{unknown} during training.
The robustness of the object detector to domain shifts is measured by the detection average precision (AP) degeneration on the test domain $\mathcal{D}_t$, compared to the model trained on the in-domain data. 

This problem setup differs from the typical unsupervised domain adaptation (UDA) setup~\cite{cai2019exploring,chen2018domain,hsu2020every,kim2019diversify,saito2019strong,zhu2019adapting}, which considers a projected adaptation between known source and target domains. In UDA, the target domain data is available for training though the labels are missing. In this work, we focus on the out-of-domain generalization setting, and we also show in the experiment sections that our approach can be extended to the UDA setting and outperform specially designed feature alignment based approaches. 

\subsection{Model Overview}
\label{sec:joint_learning}
We consider training an object detector in tandem with the auxiliary self-supervised task,
\task,  as shown in Figure~\ref{fig:arch}. \task and the object detector share the
feature extractor and are jointly trained from scratch. In contrast to pre-training
on unlabeled data using self-supervised tasks, we view the self-supervised task as regularization and train the model in a multitask manner. 

The \task head can be replaced with 
other self-supervised tasks like rotation and Jigsaw, which share the low-level image features with the object detector rather than instance features as \task does. 

\minisection{Base object detector.} We adopt the two-stage object detector, Faster R-CNN~\cite{ren2015faster}, as the base object detector, which can be replaced with other off-the-shelf detectors. 
A typical region-based object detector is composed of three components. A backbone feature extractor $\mathcal{B}$ is used to obtain image features. A set of region proposals $\mathbf{p}$ is obtained from a region proposal network (RPN).
Through ROI Align~\cite{ren2015faster}, we obtain a set of ROI features, which are then fed into the box predictors for box classification and localization.

\task uses the region features produced by
the RPN network as input while other
self-supervised tasks, such as rotation and Jigsaw, use
the image features produced by the backbone feature extractor. The objective function to train the entire model is defined as 
\begin{equation}
    \min\mathcal{L} = \min\left[\mathcal{L}_{det} + \gamma\cdot\mathcal{L}_{sup}\right], 
    \label{eq:obj}
\end{equation}
where $\gamma$ is the scaling factor when combining the detection loss $\mathcal{L}_{det}$ and
the self-supervised loss $\mathcal{L}_{sup}$. We use $\gamma=0.01$
in our experiments if not explicitly mentioned. 

\begin{figure}[t]
  \centering
  \includegraphics[width=\linewidth]{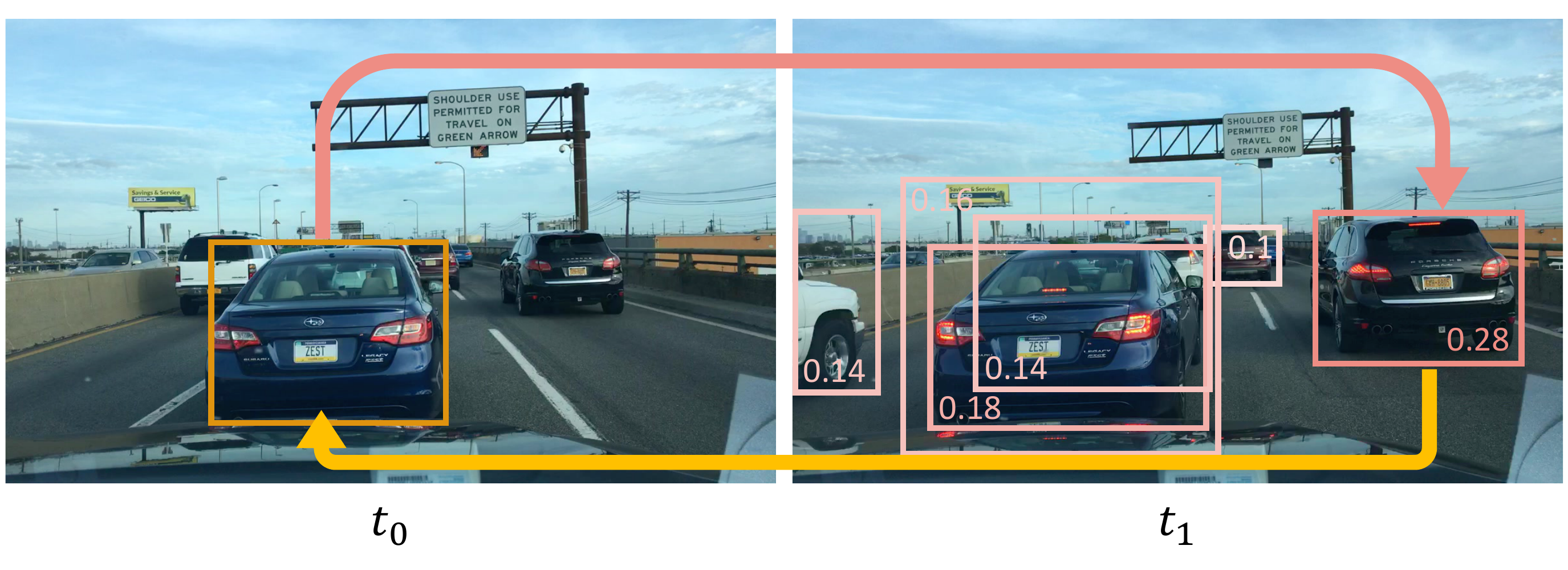}
  \vspace{-5mm}
  \caption{\small
  Formation of a time cycle.
  In the forward pass, the object in the first frame is matched to the most different target in the second frame.
  The soft target is then matched back to the original object itself in the backward pass to form a cycle. \vspace{-2mm}
  }
  \label{fig:cycle}
\end{figure}

\subsection{Instance-Level Cycle Confusion}
\label{sec:cycle_inconsist}
In this section, we dive into the details of the instance-level cycle confusion (\task) task. 

\minisection{Form a time cycle.} \task operates
on the region features produced by the region proposal
network (RPN). For a pair of images at time stamps 
$t_0$ and $t_1$, we collect a set of proposals for each object
with objectiveness score above a threshold $S$.
As shown in
Figure~\ref{fig:cycle}, we consider a matching cycle in time. 
For an instance $I_0$ at $t_0$, in the forward pass from $t_0$ to
$t_1$, we find a matching target $I_1$ at $t_1$. In the backward pass from $t_1$ to $t_0$, $I_1$ should match back to original instance $I_0$ at $t_0$ as the self-supervision signal. 

\minisection{Find the soft target.} To determine the matching objects across frames, we adopt a similarity measurement using the $L_2$ distance between two instance features. That is, 
\begin{equation}
    s_{i,j} = ||\mathbf{u}_i^0 - \mathbf{v}_j^1||_2^2,
    \label{eq:sim}
\end{equation}
where $\mathbf{u}_i^0$ is the i-th instance feature at $t_0$ and $\mathbf{v}_j^1$ is the j-th instance feature at $t_1$. $\mathbf{u}_i^0$ and $\mathbf{v}_j^1$ are single-dimension vectors obtained by a small encoder network using the region of interest (ROI) features as input. In our experiments, the encoder consists two convolutional layers with kernel size of $3\times 3$ and an average pooling layer.
The larger $s_{i,j}$ is, the more different the instances features are. 

Following the practice in the cycle consistency literature~\cite{dwibedi2019temporal,wang2019learning}, we consider a \emph{soft matching} target, which is a weighted average of the instance features at $t_1$ to avoid the instability caused by matching to a single instance feature. The weights $\alpha$ are defined as the normalized exponential of the similarity scores as follows. 
\begin{equation}
\hat{\mathbf{v}}^i = \sum_{j=1}^{N_1} \alpha_{i,j}\mathbf{v}_j^1, \:\:\: \alpha_{i,j} = \frac{e^{s_{i,j}}}{\sum_{k=1}^{N_1} e^{s_{i, k}}}, 
\label{eq:soft_target}
\end{equation}
where $\hat{\mathbf{v}}^i$ is the soft target of the i-th instance at $t_0$ and $N_1$ is the number of object proposals at $t_1$.

\begin{figure}[t]
  \centering
  \begin{subfigure}[b]{0.16\textwidth}
    \centering
    \includegraphics[width=0.9\linewidth]{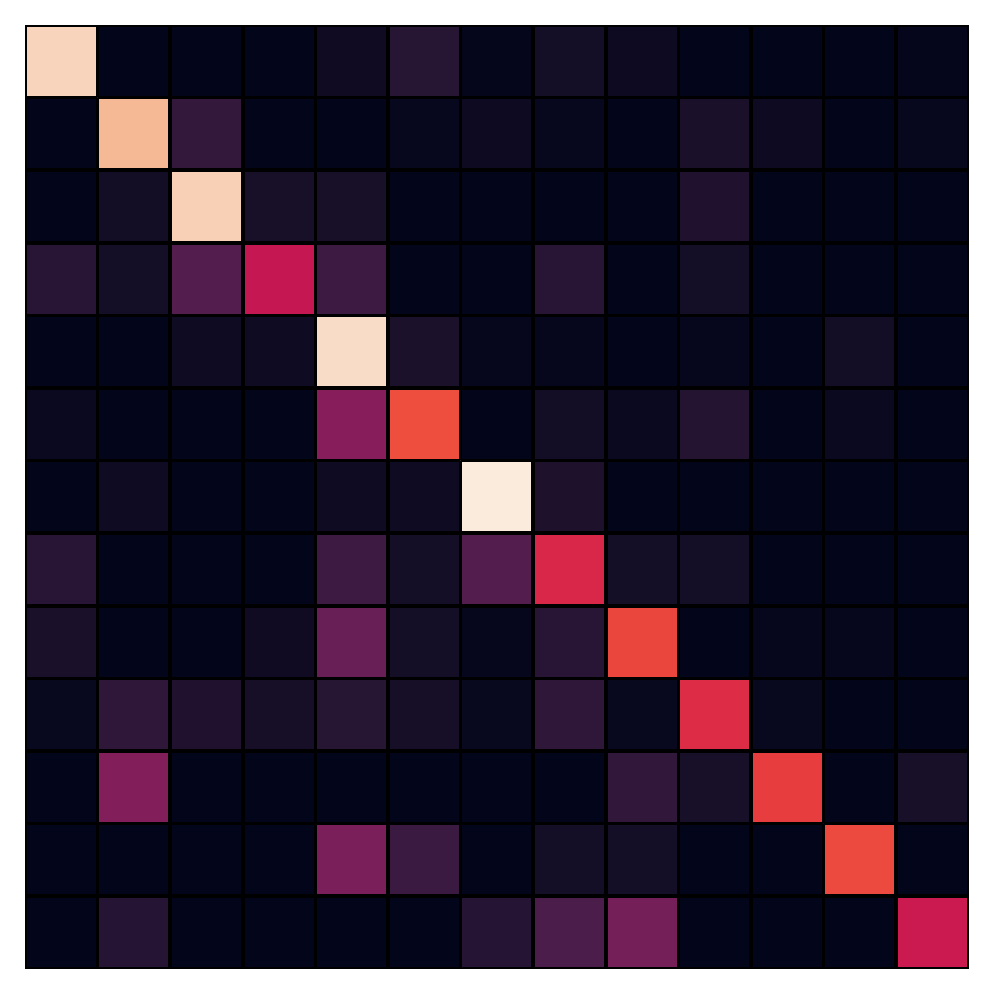}
    \vspace{-0.9mm}
    \caption{Nearest Neighbors}
  \end{subfigure}
  \begin{subfigure}[b]{0.20\textwidth}
    \centering
    \hspace*{5mm}
    \includegraphics[width=0.9\linewidth]{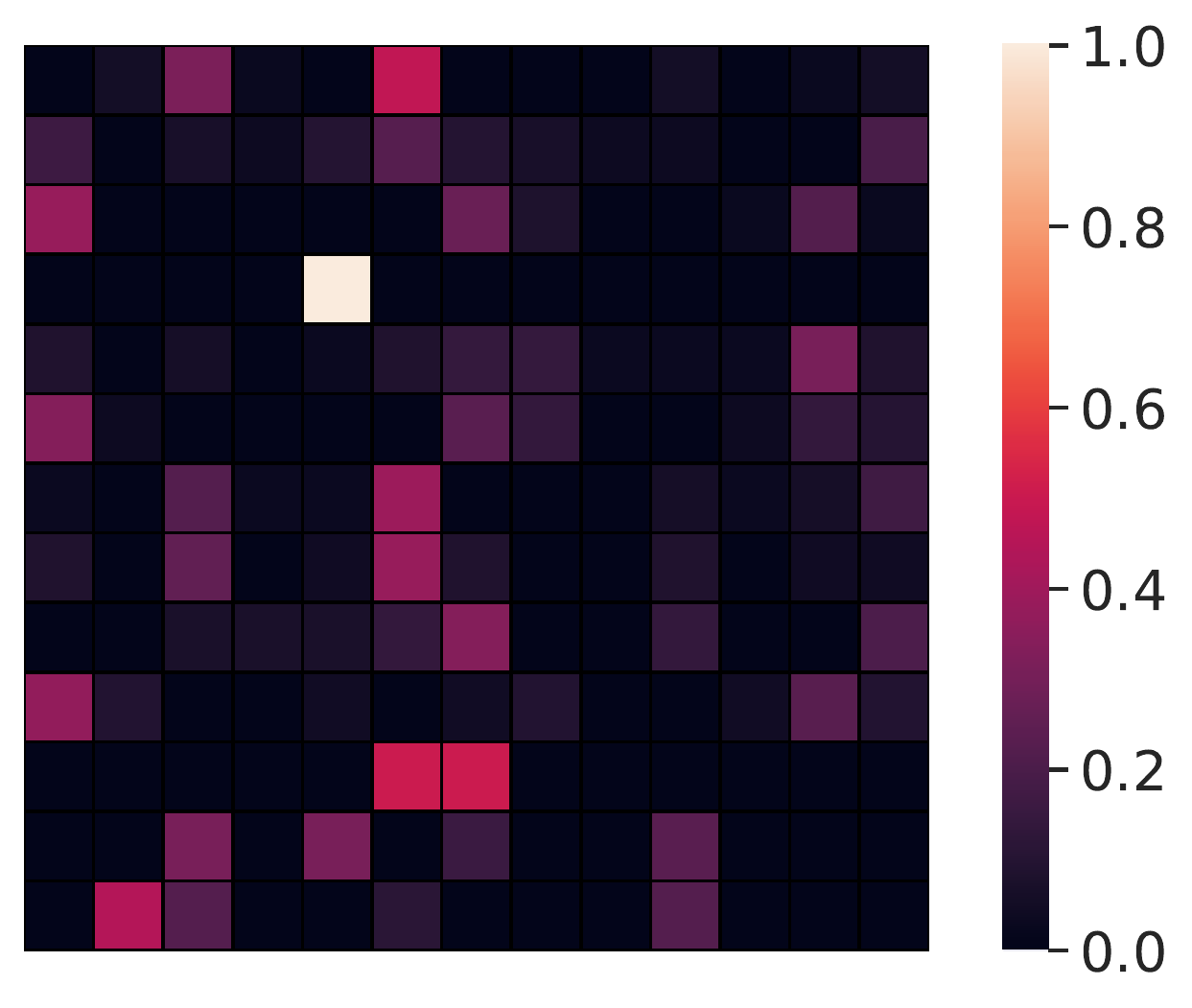}
    \vspace{-6mm}
    \caption{\task}
  \end{subfigure}
  \vspace{-1.4mm}
  \caption{\small
  Forward matching probabilities of instance features.
  Each row represents a different instance.
  The matching probability distribution of \task is more flat and scattered, which means its entropy value is higher than nearest neighbors'.
  }
  \vspace{-1mm}
  \label{fig:match_prob}
\end{figure}

Since we have a fixed confidence score threshold $S$ to select object proposals, the size of $N_1$ increases and the matching task becomes more challenging as the object detector is better trained to produce high quality object proposals.  Naturally, \task forms a curriculum training scheme. 

\minisection{Match back to self.} Similarly, the backward pass requires the soft target at $t_1$ to match to the original object at $t_0$. We view this backward matching as a classification task, where a cross-entropy loss is adopted as the training objective.
We use similarity scores $\mathbf{s}^i\in\mathbb{R}^{N_0}$ defined as
\begin{equation}
    \mathbf{s}^i = [s_{1, i}, ... s_{k, i}, ..., s_{N_0, i}],  \:\:\: s_{k, i} = ||\mathbf{u}_k^0 - \hat{\mathbf{v}}^i||_2^2.
\label{eq:back_sim}
\end{equation}
The similarity scores are used as the logits and the loss function is defined as 
\begin{equation}
    \mathcal{L}_{sup}(\mathbf{u}, \hat{\mathbf{v}}^i) = -\sum_k^{N_0}\mathbb{1}(k=i)\log\left (\texttt{softmax}(\mathbf{s}^i)_k\right).
\end{equation}
Taking all the instance features $\mathbf{u}_i^0$ into account, the overall cycle confusion loss is defined as 
\begin{equation}
    \mathcal{L}_{sup}(\mathbf{u}, \hat{\mathbf{v}}) = \frac{1}{N_0}\sum_{i=1}^{N_0}\mathcal{L}_{sup}(\mathbf{u}, \hat{\mathbf{v}}^i). 
\end{equation}
We train the object detector and the \task task jointly by combining the detection loss and the self-supervised task loss (Equation~\ref{eq:obj}).

\begin{figure*}[tp]
    \centering
    \begin{subfigure}[b]{0.22\textwidth}
        \centering
        \includegraphics[width=\textwidth]{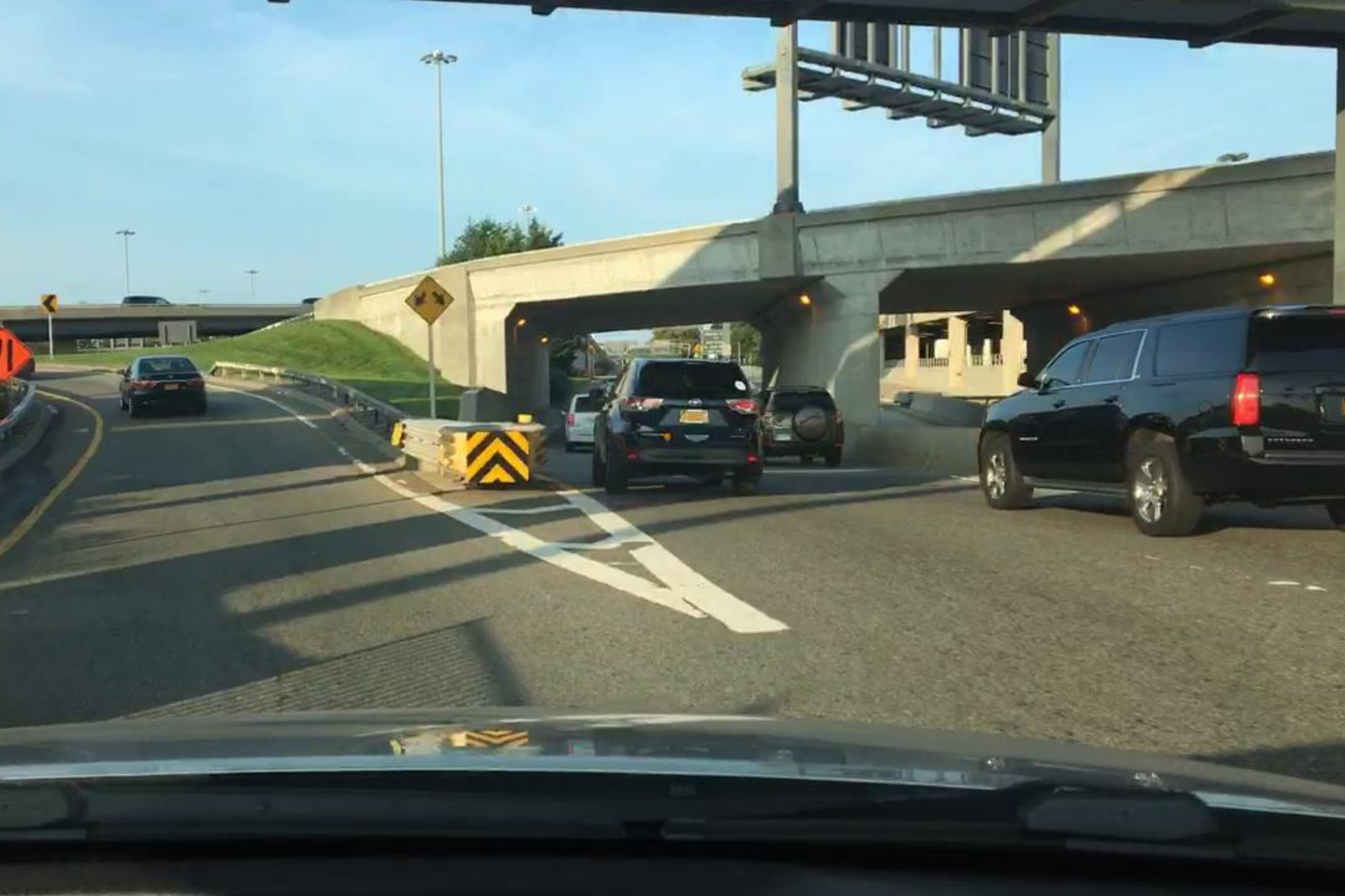}
        \caption{BDD100K Daytime}
    \end{subfigure}
    \begin{subfigure}[b]{0.22\textwidth}
        \centering
        \includegraphics[width=\textwidth]{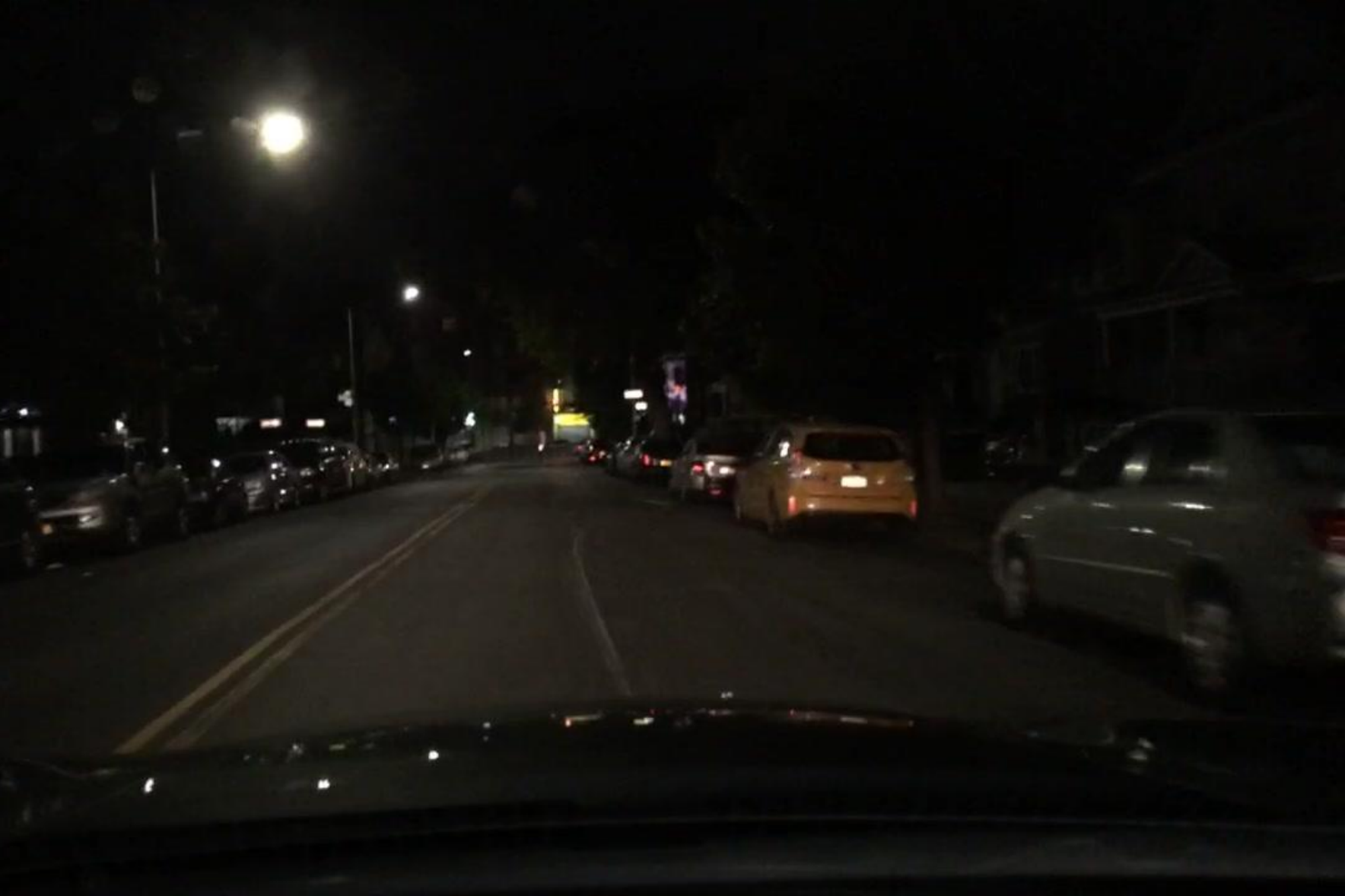}
        \caption{BDD100K Night}
    \end{subfigure}
    \begin{subfigure}[b]{0.22\textwidth}
        \centering
        \includegraphics[width=\textwidth]{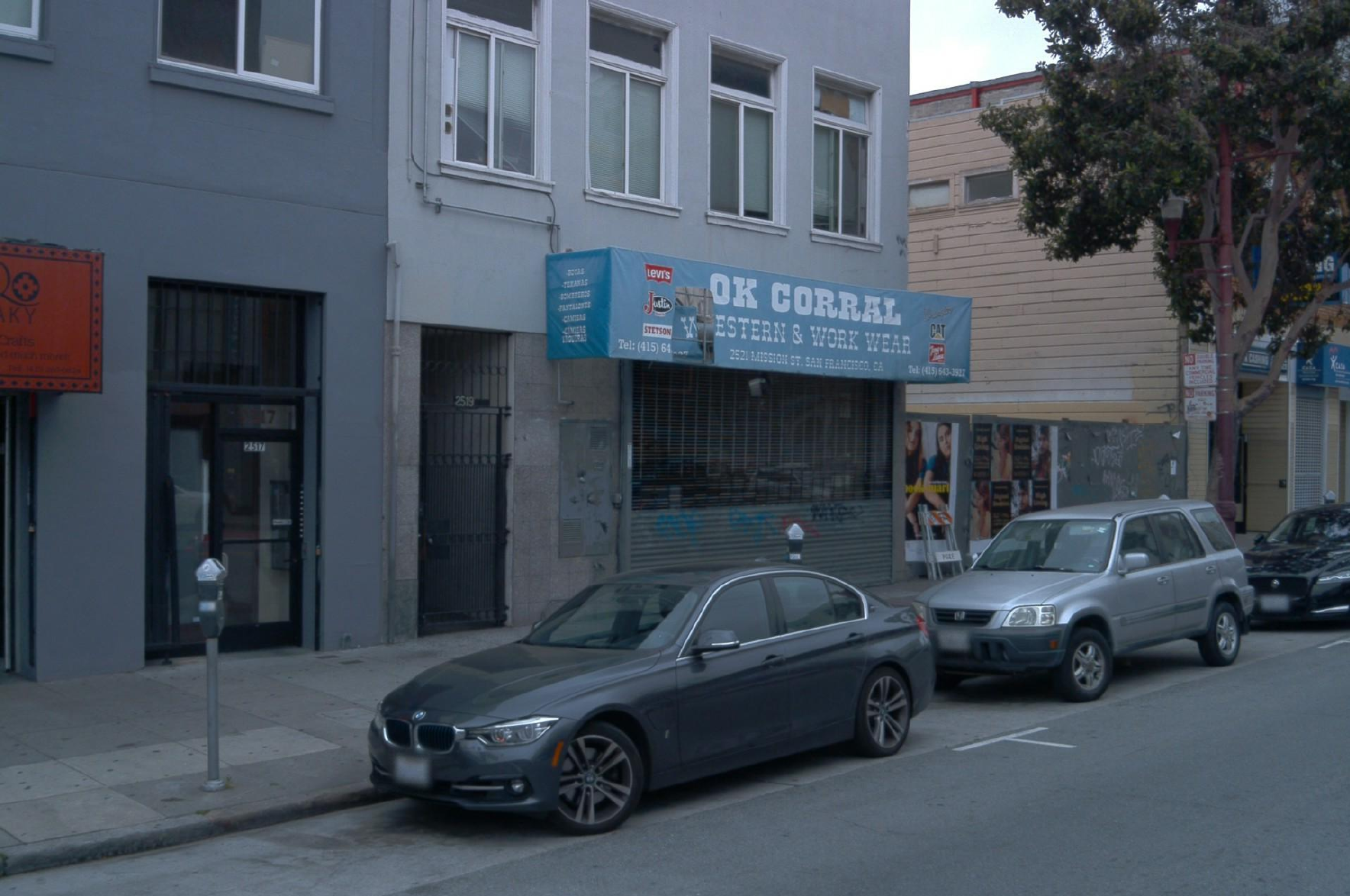}
        \caption{Waymo Front Left}
    \end{subfigure}
    \begin{subfigure}[b]{0.22\textwidth}
        \centering
        \includegraphics[width=\textwidth]{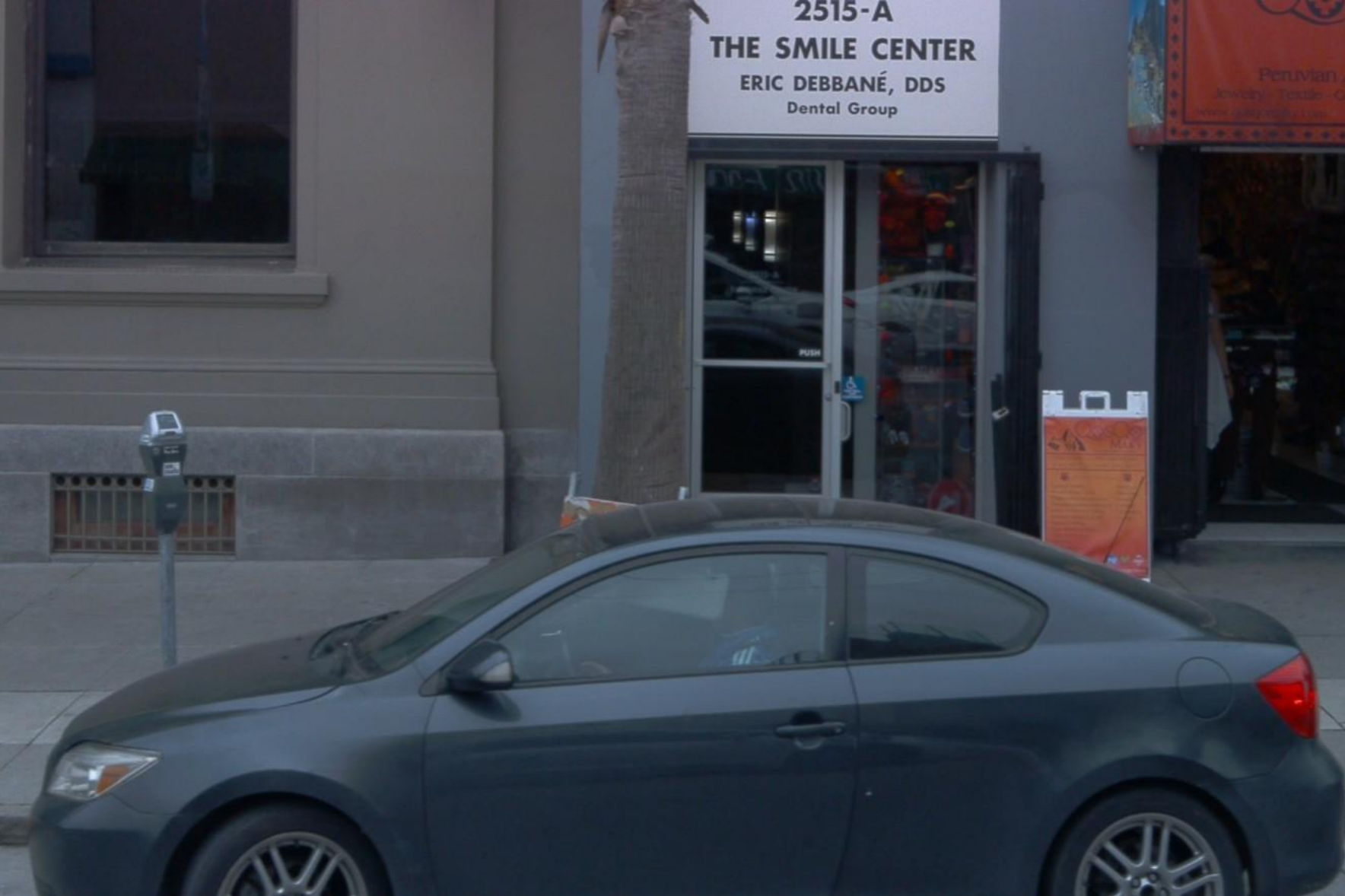}
        \caption{Waymo Side Left}
    \end{subfigure}
    \vspace{-2mm}
    \caption{\small Training samples from BDD100K and Waymo Open datasets. The Waymo open dataset has different camera angles as BDD100K.\vspace{-3mm}}
    \label{fig:samples}
\end{figure*}

\minisection{Maximum entropy interpretation.}
The matching procedure in \task can be interpreted as a maximum entropy regularization.
In Figure~\ref{fig:match_prob}, we visualize the matching probability distribution in the forward pass of \task and compare it with nearest neighbors on the instance features.
We can see from the figure, nearest neighbors tends to match the identical objects across the frames since the diagonal instances have much higher matching probability.
In contrast, the matching probability distribution in \task is flat and scattered, whose entropy value is higher than nearest neighbors'.
In technical terms, \task can be viewed as maximizing the entropy of the probability distribution of the weights $\alpha$, 
\begin{equation}
    \max H (\alpha) = -\frac{1}{N_0}\sum_{i=0}^{N_0}\sum_{j=1}^{N_1} \alpha_{i,j} \log(\alpha_{i,j}).
\end{equation}
The principle of maximum entropy~\cite{jaynes1957information,jaynes1957information2} indicates the probability distribution with the highest entropy is the one that best represents the current state of knowledge in the context of precisely stated prior data,
which in the recent literature~\cite{eysenbach2021maximum,rawlik2012stochastic,todorov2006linearly,ziebart2010modeling} shows can improve model robustness. 

\section{Out-of-Domain Evaluation}

In this section, 
we describe the out-of-domain evaluation benchmark for object detectors using large-scale driving video datasets BDD100K~\cite{bdd100k} and the Waymo Open Dataset~\cite{sun2019scalability} in Section~\ref{sec:benchmark}. We show that \task outperforms other commonly used self-supervised tasks when evaluated on the unseen test domains in Section~\ref{sec:ood_eval}.  Visualization and analysis are presented in Section~\ref{sec:ablation}.

\subsection{Benchmark Construction}
\label{sec:benchmark}
\begin{table}[tp]
\centering
\caption{\small Dataset statistics of BDD100K and Waymo Open data. \vspace{-2mm}}
\label{table:data_stats}
\adjustbox{width=\linewidth}{
\begin{tabular}{@{}lccccc@{}} \toprule
Dataset & Split & seq. & frames/seq. & boxes & classes \\\midrule
\multirow{2}{*}{BDD100K Daytime} & Train & 757 & 204 & 1.82M & 8 \\ 
 & Val & 108 & 204 & 287K & 8 \\\midrule
 \multirow{2}{*}{BDD100K Night} & Train & 564 & 204 & 895K & 8 \\
 & Val & 71 & 204 & 137K & 8 \\\midrule
\multirow{2}{*}{Waymo Open Data} & Train & 798 & 199 & 3.64M & 3 \\
 & Val & 202 & 199 & 886K & 3 \\
\bottomrule
\end{tabular}}
\vspace{-2mm}
\end{table}

\minisection{Datasets.} For BDD100K, we divide the dataset into non-overlapping daytime and night splits to create a domain gap.
We denote the splits as \emph{BDD100K Daytime} and \emph{BDD100K Night}. For the Waymo Open data, we split the dataset based on five different angles including front, front left, front right, side left, and side right.
The front camera angle is consistent with the camera angle in the BDD100K dataset, while the other camera angles are not. The statistics of all the datasets are provided in Table~\ref{table:data_stats} and some example training samples are provided in Figure~\ref{fig:samples}.

\minisection{Evaluation settings.} We consider two out-of-domain evaluation scenarios: (1) \emph{Domain Shift by Time of Day}, where the model is trained and evaluated on different time of day. (2) \emph{Cross-camera domain shift}, where the model is trained and evaluated under different camera views. Since the Waymo Open data and BDD100K are collected from different sensors, the dataset distribution shift is more severe than the first scenario. 

\minisection{Baselines.}
We consider three self-supervised tasks and compare them with the proposed Faster R-CNN w/ \task. 
\setlist{nolistsep}
\begin{itemize}[noitemsep,leftmargin=*]
    \item Faster R-CNN w/ Rotation~\cite{gidaris2018unsupervised}. 
    We jointly train Faster R-CNN with the image-level rotation task, where each image is rotated, and the detector additionally has to predict the angle of rotation.
    \item Faster R-CNN w/ Jigsaw~\cite{noroozi2016unsupervised}. For Jigsaw, a 2x2 grid is sampled from each image and shuffled, and the detector has to predict the permutation of the tiles. 
    \item Faster R-CNN w/ Cycle Consistency. The instance-level cycle consistency task is to find the nearest neighbors across the consecutive frames. This is a baseline method we adapted from the existing literature~\cite{dwibedi2019temporal}, where we replace the random patches with the instance features produced by the object detector. 
\end{itemize}

\minisection{Implementation details.}
We conduct the experiments using PyTorch 1.6.0~\cite{NEURIPS2019_9015}
with Detectron2~\cite{wu2019detectron2} library.
For all experiments, we use Faster R-CNN~\cite{ren2015faster} as our base detector and Resnet-50~\cite{he2016deep} with a Feature Pyramid Network~\cite{lin2016feature} as the backbone.
All models are trained on 8 GPUs using SGD with a mini-batch size of 16, momentum of 0.9, and weight decay of 0.0001.  We set the object score threshold $S=0.8$ in our experiments if not mentioned. 

\begin{table*}[htp]
\centering
\caption{\small BDD100K Daytime and Night. \task outperforms other self-supervised tasks in both settings.\vspace{-3mm}}
\adjustbox{width=.9\linewidth}{
\begin{tabular}{l|cccccc|cccccc}
\toprule
& \multicolumn{6}{c|}{BDD100K Daytime $\rightarrow$ Night}    & \multicolumn{6}{c}{BDD100K Night $\rightarrow$ Daytime}\\ 
\midrule\cmidrule{2-13}
Model  & AP  & AP50 & AP75  & APs   & APm   & APl  & AP & AP50 & AP75 & APs & APm & APl \\
\midrule
Faster R-CNN        & 17.84 & 31.35 & 17.68 & 4.92 & 16.15 & 35.56 & 19.14 & 33.04 & 19.16 & 5.38 & 21.42 & \textbf{40.34}  \\ \midrule
+ Rot  & 18.58 & 32.95 & 18.15 & 5.16 & 16.93 & \textbf{36.00} & 19.07 & 33.25 & 18.83 & 5.53 & 21.32 & 40.06 \\
+ Jigsaw  &  17.47 & 31.22 & 16.81 & 5.08 & 15.80 & 33.84 & 19.22 & 33.87 & 18.71 & 5.67 & 22.35 & 38.57 \\
+ Cycle Consistency & 18.35 & 32.44 & 18.07 & 5.04 & 17.07 & 34.85 & 18.89 & 33.50 & 18.31 & 5.82 & 21.01          & 39.13 \\ \midrule
+ Cycle Confusion (Ours)   & \textbf{19.09} & \textbf{33.58} & \textbf{19.14} & \textbf{5.70} & \textbf{17.68} & 35.86 & \textbf{19.57} & \textbf{34.34} & \textbf{19.26} & \textbf{6.06} & \textbf{22.55} & 38.95 \\ 
\bottomrule
\end{tabular}}
\label{tbl:bdd}
\end{table*}

\begin{table*}[htp]
\centering
\caption{\small Waymo to BDD100K. The domain gap is due to the changing camera angles. \task outperforms other self-supervised tasks in both settings by a relatively large margin.\vspace{-3mm}}
\label{tab:waymo2bdd}
\adjustbox{width=.9\linewidth}{
\begin{tabular}{l|cccccc|cccccc}
\toprule
 & \multicolumn{6}{c|}{Waymo Front Left $\rightarrow$ BDD100K Night} & \multicolumn{6}{c}{Waymo Front Right $\rightarrow$ BDD100K Night} \\ \cmidrule{2-13}
Model & AP & AP50 & AP75 & APs & APm & APl & AP & AP50 & AP75 & APs & APm & APl \\ \midrule
Faster R-CNN &  10.07 & 19.62 & 9.05 & 2.67 & 10.81 & 18.62 & 8.65 & 17.26 & 7.49 & 1.76 & 8.29 & 19.99 \\\midrule
+ Rot & 11.34 & 23.12 & 9.65 & \textbf{3.53} & 11.73 & 21.60 & 9.25 & 18.48 & 8.08 & 1.85 & 8.71 & \textbf{21.08} \\
+ Jigsaw &  9.86 & 19.93 & 8.40 & 2.77 & 10.53 & 18.82 & 8.34 & 16.58 & 7.26 & 1.61 & 8.01 & 18.09 \\
+ Cycle Consistency &  11.55 & 23.44 & 10.00 & 2.96 & 12.19 & 21.99 &  9.11 & 17.92 & 7.98 & 1.78 & 9.36 & 19.18 \\ \midrule
+ Cycle Confusion (Ours) &  \textbf{12.27} & \textbf{26.01} & \textbf{10.24} & 3.44 & \textbf{12.22} & \textbf{23.56} &  \textbf{9.99} & \textbf{20.58} & \textbf{8.30} & \textbf{2.18} & \textbf{10.25} & 20.54 \\ \bottomrule
\end{tabular}}
\label{tbl:waymo}
\end{table*}

\subsection{Evaluation Results}
\label{sec:ood_eval}
\minisection{Domain Shift by Time of Day.}
We first evaluate on the large scale BDD100K dataset. We construct two settings on this dataset, BDD100K Daytime $\rightarrow$ Night and BDD100K Night $\rightarrow$ Daytime. For each setting, we train our detector on one split and evaluate on the other split.
The results for BDD100K are shown in Table~\ref{tbl:bdd}.

In both settings, our proposed \task method is able to significantly improve the base detector's performance.
In particular, on BDD100K Daytime $\rightarrow$ Night, \task can achieve around 2 points improvement in AP50 and 1.5 points improvement in AP over the base detector.
Compared to other self-supervised tasks, \task can also lead to better performance across both settings.
\task is able to outperform the cycle consistency baseline consistently, indicating that \task helps the detector generalize better compared to cycle consistency.

\minisection{Cross-camera Domain Shift.}
We construct two settings on this dataset, Waymo Front Left $\rightarrow$ BDD100K Night and Waymo Front Right $\rightarrow$ BDD100K Night.
Waymo Front Left and Waymo Front Right consist of images taken from the front left and the front right cameras, respectively.
We train our detector on the Waymo data and then evaluate on BDD100K Night.
Since there are changes in the view angle, the domain gap in these two settings is larger than that in the previous settings.

The results for Waymo to BDD100K are shown in Table~\ref{tbl:waymo}.
Despite the larger domain gap, our proposed \task is still able to outperform the other methods by a relatively large margin.
On Waymo Front Right $\rightarrow$ BDD100K Night, \task can achieve around 2 points improvement in AP50 and 1 point in AP.
In both settings, \task is able to significantly improve the performance of the base detector by over 3 points in AP50.
\task also performs consistently better than the model trained with cycle consistency.

\subsection{Visualization and Analysis}
\label{sec:ablation}

\begin{figure}[tp]
    \centering
    \begin{subfigure}[b]{0.235\textwidth}
        \centering
        \includegraphics[width=\textwidth]{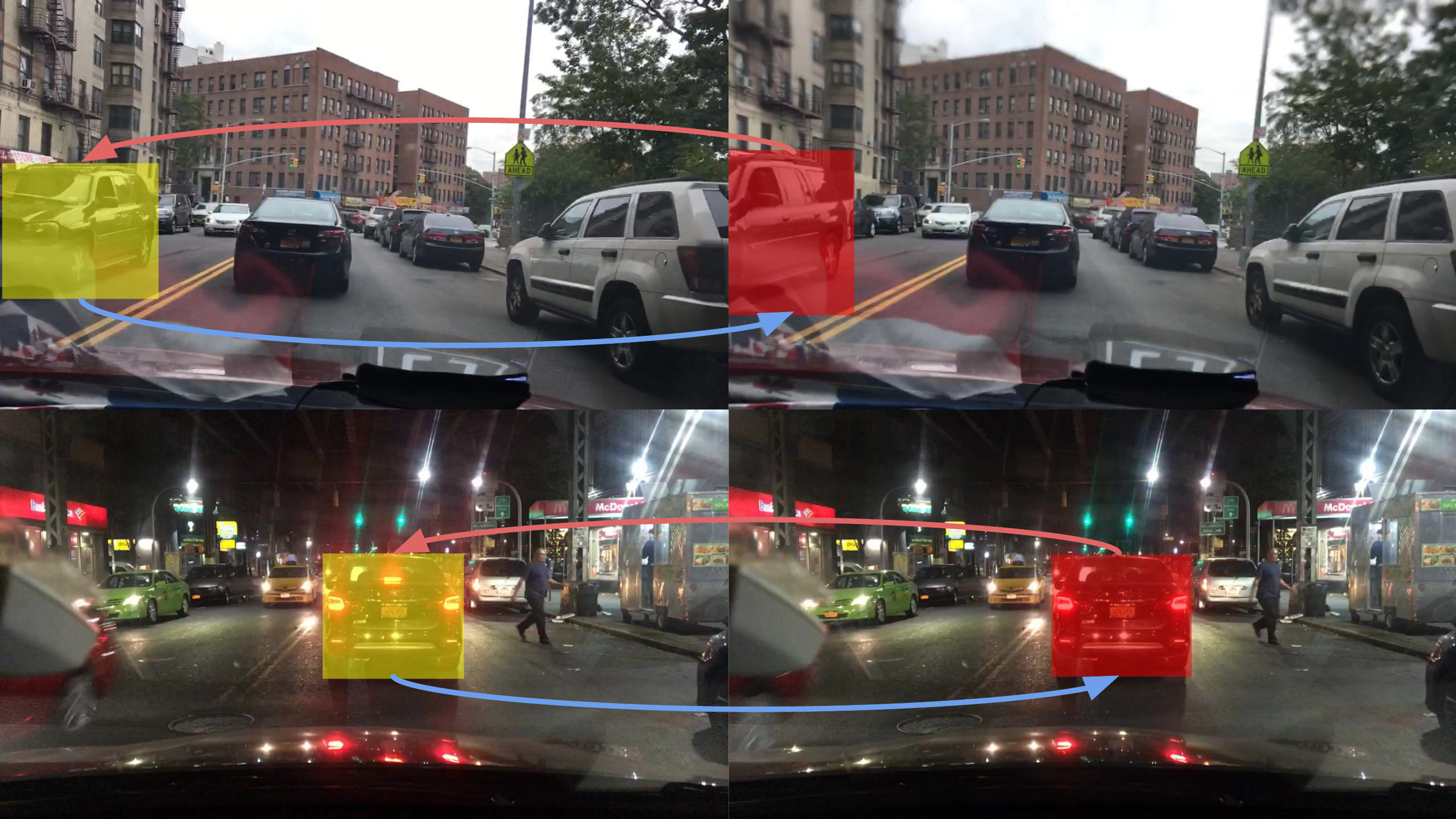}
        \caption{Cycle Consistency}
    \end{subfigure}
    \begin{subfigure}[b]{0.235\textwidth}
        \centering
        \includegraphics[width=\textwidth]{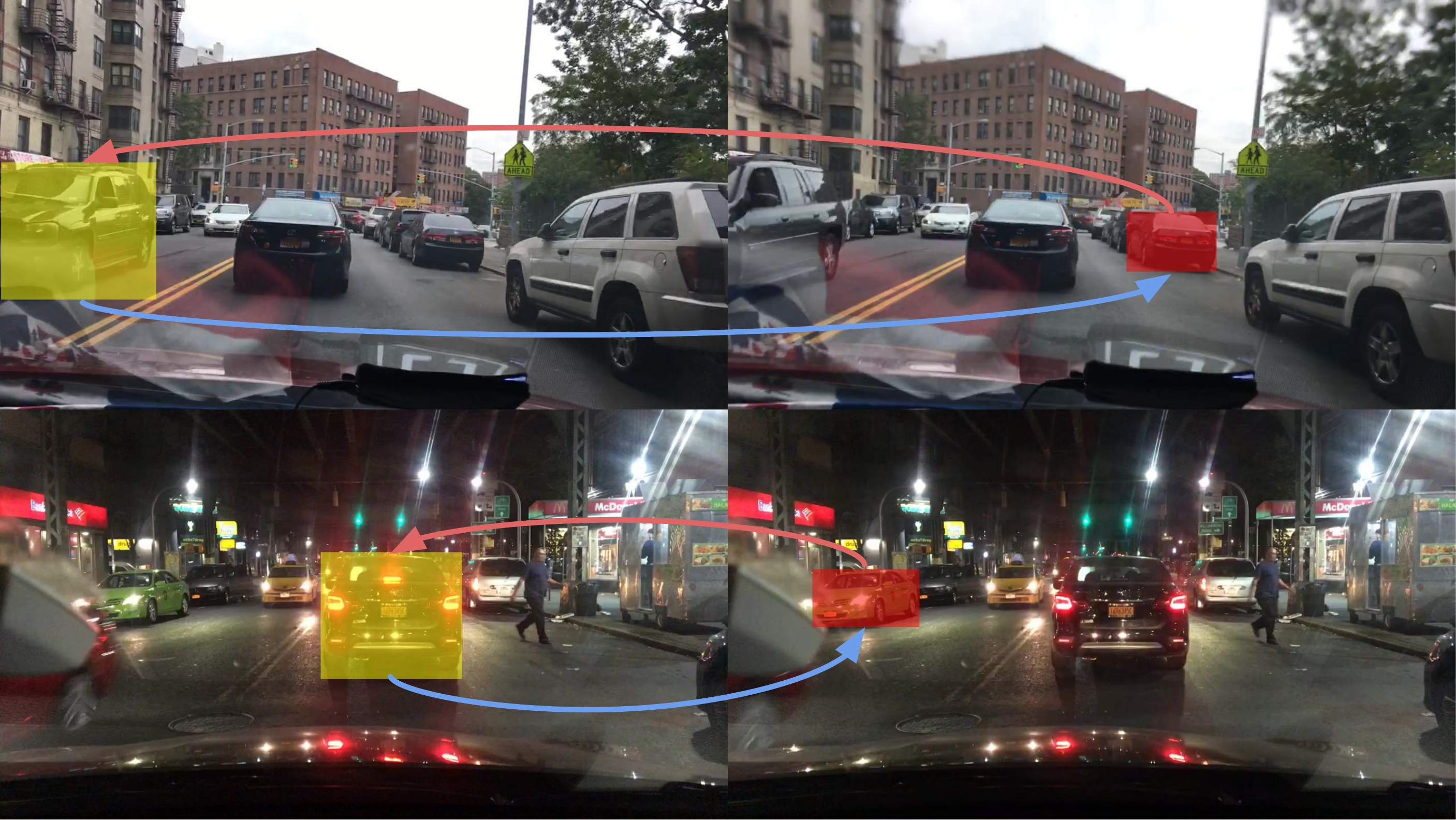}
        \caption{Cycle Confusion}
    \end{subfigure}
    \vspace{-6mm}
    \caption{\small Examples of proposal matches of the detector trained with cycle consistency and CycConf on BDD100K. When using cycle consistency, the matched proposals are similar, while CycConf matches dissimilar proposals.  \vspace{-4mm}}
    \label{fig:matching}
\end{figure}

\minisection{Proposal matching.}
We visualize proposal matches in Figure~\ref{fig:matching} to show the difference between the behavior of instance-level cycle consistency and \task.
For the former, each proposal is encouraged to be matched to its nearest neighbor in the next frame, and thus it tends to be matched to itself in the next frame.
For \task, each proposal is matched to the most different proposal in the next frame.
In the first sequence (top row), the car on the left is matched to the car on the right.
In the second sequence (bottom row), the car in the front is matched to the green car on the left.

\begin{figure*}[tp]
    \centering
    \begin{subfigure}[b]{0.24\textwidth}
        \centering
        \includegraphics[width=\textwidth]{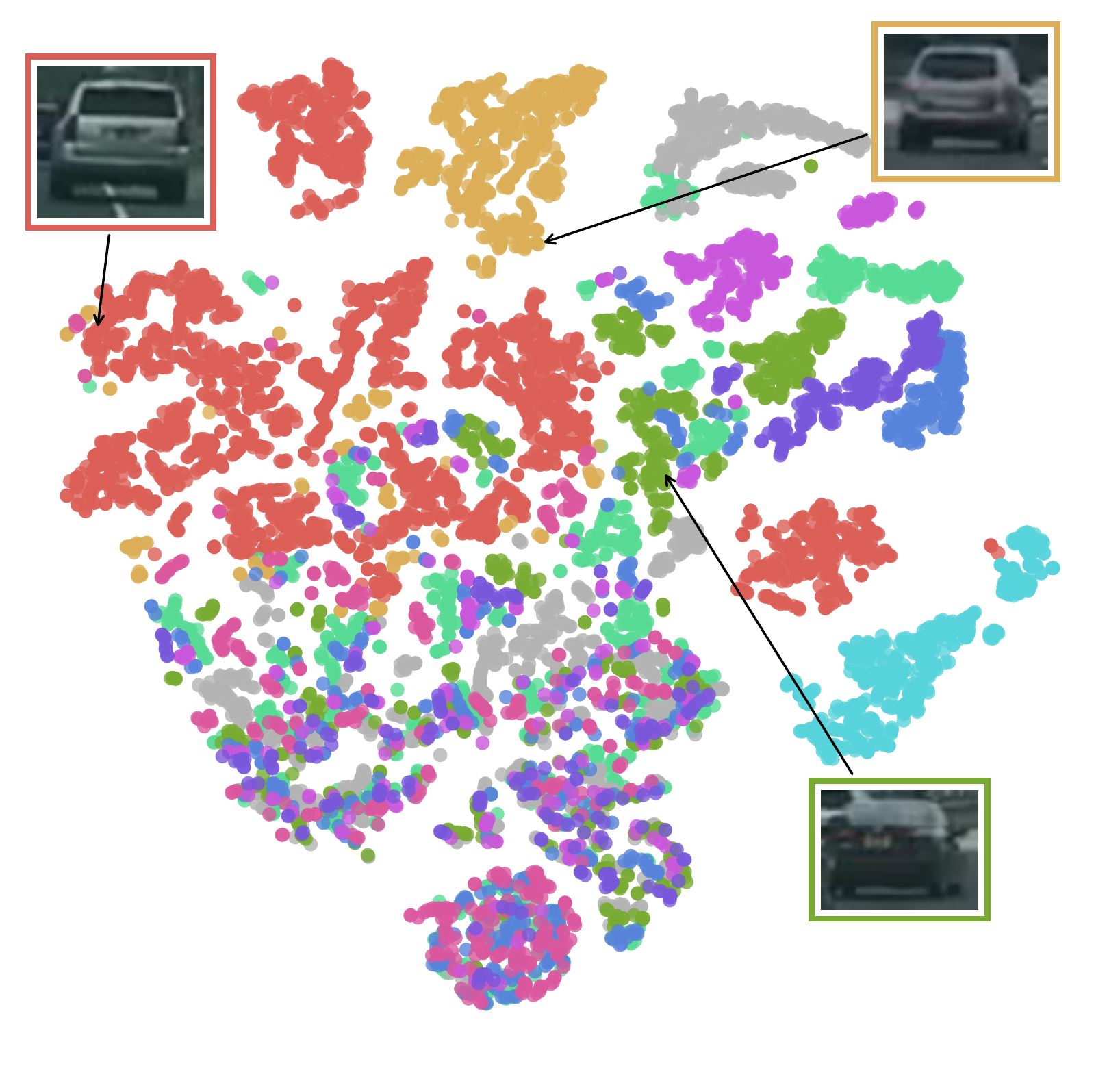}
        \caption{Cycle Consistency (Daytime)}
    \end{subfigure}
    \begin{subfigure}[b]{0.24\textwidth}
        \centering
        \includegraphics[width=\textwidth]{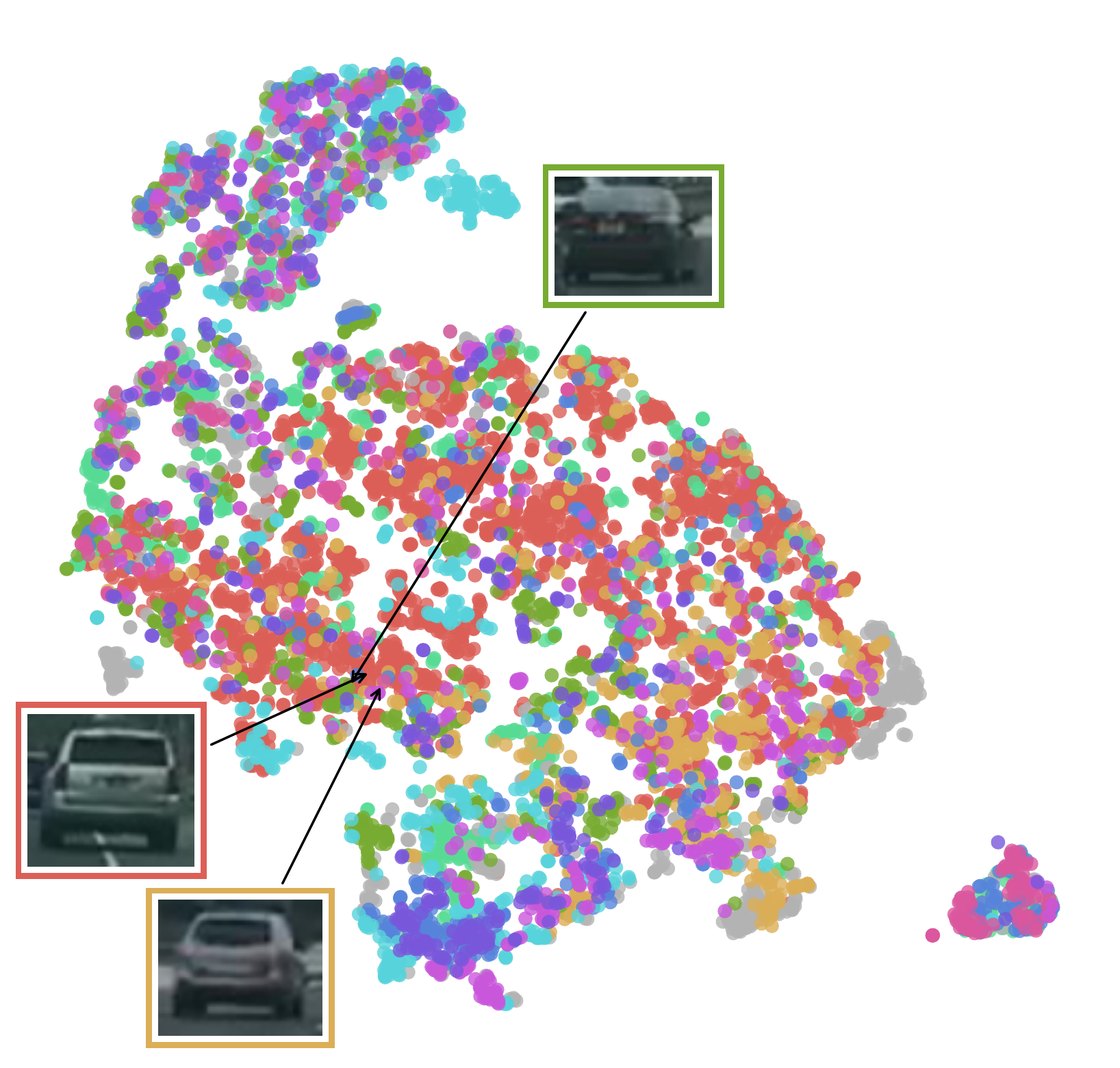}
        \caption{\task (Daytime)}
    \end{subfigure}
    \begin{subfigure}[b]{0.24\textwidth}
        \centering
        \includegraphics[width=\textwidth]{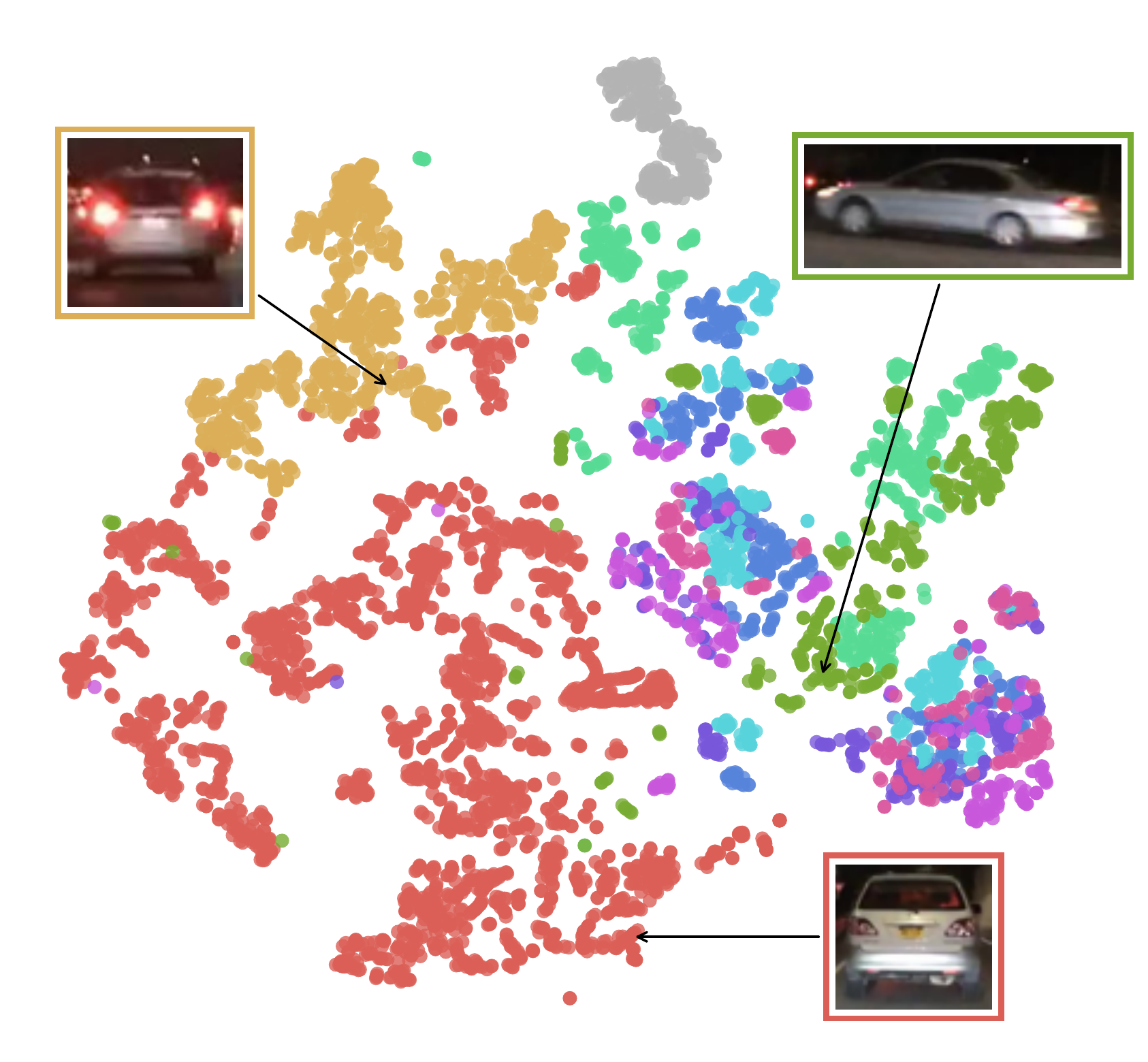}
        \caption{Cycle Consistency (Night)}
    \end{subfigure}
    \begin{subfigure}[b]{0.24\textwidth}
        \centering
        \includegraphics[width=\textwidth]{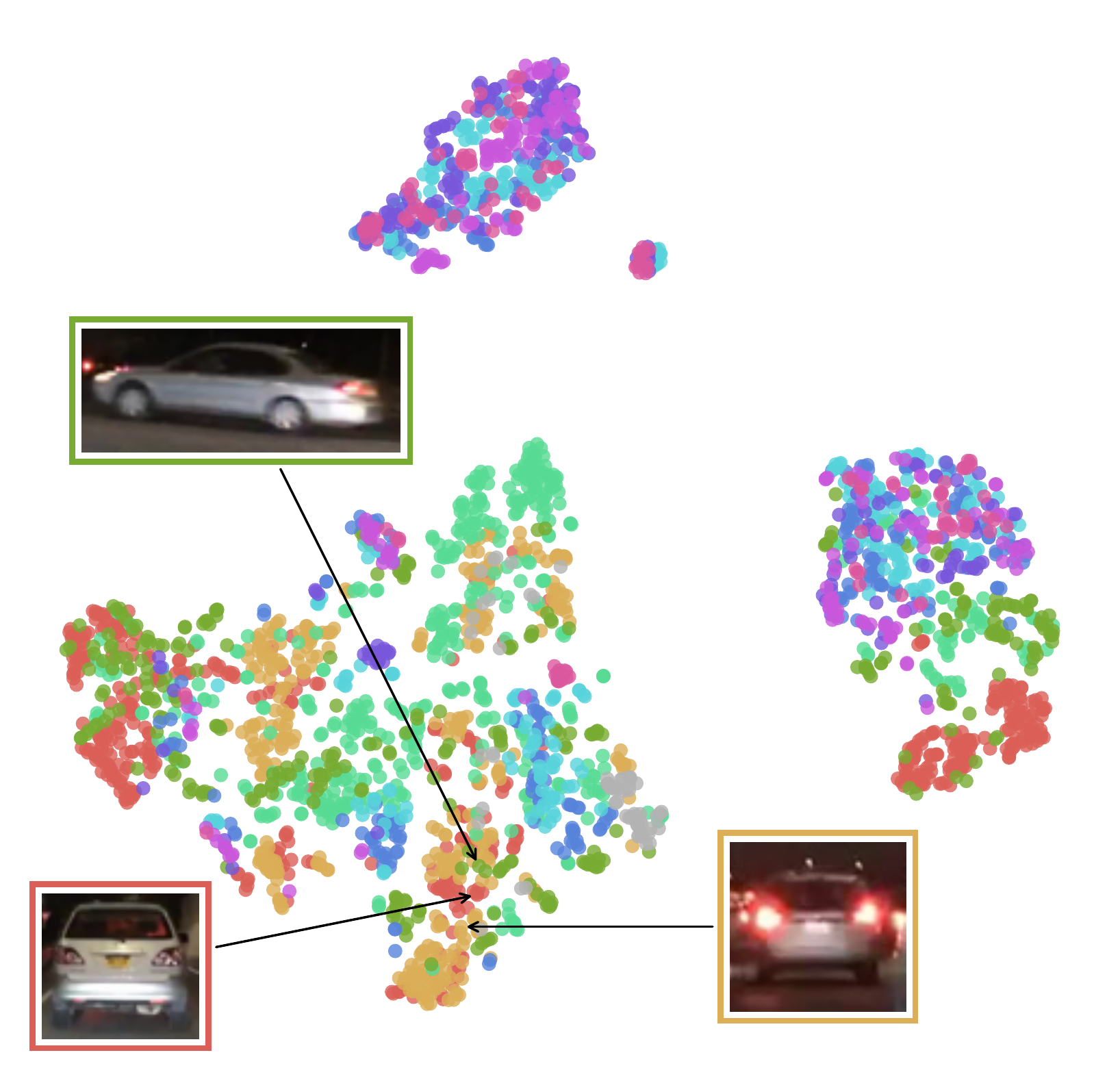}
        \caption{\task (Night)}
    \end{subfigure}
    \vspace{-1mm}
    \caption{\small t-SNE visualization of instance features from Cycle Consistency ((a) and (c)) and \task ((b) and (d)) on BDD100K Daytime $\rightarrow$ Night. Each color represents a different instance. Images of three instances are shown for each dataset as well as the position of their features in the t-SNE space. The features of different instances when using the cycle consistency task are well separated despite their visual similarity. In comparison, \task's features are more mixed, which encourages the detector to explore the latent structures of each instance. \vspace{-5mm}}
    \label{fig:tsne}
\end{figure*}

\minisection{Instance features.}
We use t-SNE to visualize the instance features using cycle consistency and \task on BDD100K Daytime $\rightarrow$ Night in Figure~\ref{fig:tsne}.
We extract features of proposals that are matched to ten car instances in a sequence of images from both datasets.
Each color represents features of a different instance.
We additionally show example images of three instances and the position of their features in the t-SNE space.
On both datasets, the instance features learned with cycle consistency are well separated despite the visual similarity between the different instances.
On the other hand, \task maps similar objects to features that are more mixed together.
This makes it more difficult for the detector to identify the instance identities and encourages the detector to explore the latent structures of each instance.

\section{Unsupervised Domain Adaptation}
\label{sec:uda}
In this section, we extend the study to typical unsupervised domain adaptation (UDA) benchmarks, a closely related setting, to test the out-of-domain generalization.
To the best of our knowledge, existing UDA methods rely on the availability of a handful of images from the target domain, and the domain adaptive object detectors
~\cite{cai2019exploring,chen2018domain,hsu2020every,kim2019diversify,saito2019strong,zhu2019adapting} are mostly evaluated on static images and not videos.
To conduct a fair comparison,
we train the object detector jointly with the self-supervised tasks on existing UDA benchmarks with static images.
The self-supervised task used in this section is image rotation, as \task is designed to operate on videos.
The key message in these experiments is that the adoption of the self-supervised
tasks can be a general solution to multiple evaluation benchmarks, whether out-of-domain generalization or typical UDA benchmarks.
As shown in Section~\ref{sec:uda_res}, a joint training approach has better accuracy than the previous feature alignment based approaches
with much less training cost.
We hope this finding can motivate future algorithm designs.

\minisection{Weather adaptation.} We first adapt the model from Cityscapes~\cite{cordts2016cityscapes} to Foggy Cityscapes~\cite{sakaridis2018semantic}. The Cityscapes dataset is a driving 
scene dataset, containing 2,975 and 500 images in the training and validation 
set, respectively, while the Foggy Cityscapes data is created by adding synthetic fog to Cityscapes. There are eight classes: \emph{person, rider, car, truck, bus, train,
motorcycle and bicycle} in both dataset. 

\minisection{Synthetic-to-real.} Sim10K~\cite{johnson2017driving} contains 10K
synthesized images with bounding box annotations. We use images of Sim10K as the source domain and adapt the model to the Cityscapes dataset. Following previous works~\cite{chen2018domain,hsu2020every,zhu2019adapting}, only the \emph{car} class is considered. 

\minisection{Implementation details.} 
The base object detector is Faster RCNN~\cite{ren2015faster} with
FPN~\cite{lin2017feature} and uses VGG-16~\cite{simonyan2014very} and
ResNet-101~\cite{he2016deep} as the backbones. 
We use rotation~\cite{gidaris2018unsupervised} as the
self-supervised task and perform image-level joint training with 
the detector. We randomly crop the input image to a 224$\times$224
patch and select a rotation angle from 0\degree, 90\degree, 180\degree\xspace and 270\degree\xspace to construct the input and label for the self-supervised task. We jointly
optimize the model with the detection loss and the self-supervised loss on the source domain data and only with the self-supervised loss on the target domain data. We set the loss scale $\lambda$ of the rotation task to 0.5.
We use 8 GPUs for training, a batch size of 32, and the
base learning rate of 0.01. We train the model for 10K iterations in
total and divide the learning rate by 10 at 6K and 8K iterations.
For evaluation metrics, we report the mean average precision (mAP) with an intersection of union (IoU) threshold of 0.5.

\begin{table*}[ht]
\centering
\small{
\setlength{\tabcolsep}{1em}
\caption{\small Adapt from Cityscapes to Foggy Cityscapes. We report the AP50 of 8 categories and the mean AP across all classes. We adopt both VGG-16 and ResNet-101 as the backbone. Despite its simplicity, \model w/ Rot outperforms the previous methods by a large margin.\vspace{-3mm}}
\label{tab:main_city2fog}
\adjustbox{width=\textwidth}{
\begin{tabular}{lcccccccccl}
\toprule
\multicolumn{11}{c}{Cityscapes$\rightarrow$Foggy Cityscapes} \\
\midrule
Method & Backbone & person & rider & car & truck & bus & train &motorcycle &bicycle & mAP50 \\
\midrule
Faster R-CNN (source only)~\cite{chen2018domain} & \multirow{8}{*}{VGG-16} & 17.8 & 23.6 & 27.1 & 11.9& 23.8& 9.1& 14.4& 22.8 & 18.8  \\
DAF~\cite{chen2018domain} && 25.0 & 31.0 & 40.5 & 22.1 & 35.3 & 20.2 & 20.0& 27.1& 27.6 (+8.8) \\
MAF~\cite{he2019multi} && 28.2 & 39.5 & 43.9 & 23.8 & 39.9 & \textbf{33.3} & 29.2 & 33.9 & 34.0 (+15.2) \\
SW-DA~\cite{saito2019strong} && 29.9 & 42.3 & 43.5 & 24.5 & 36.2 & 32.6 & \textbf{30.0} & 35.3 & 34.3 (+15.5)\\
DAM~\cite{kim2019diversify} && 30.8 & 40.5 & 44.3 & \textbf{27.2} & 38.4 & 34.5 & 28.4 & 32.2 & 34.6 (+15.8)\\
EPM~\cite{hsu2020every} && 41.9 & 38.7 & 56.7 & 22.6 & 41.5 & 26.8 & 24.6 & 35.5 & 36.0 (+17.2)\\
\textcolor{gray}{Faster R-CNN (oracle)} && \textcolor{gray}{47.4} & \textcolor{gray}{40.8} & \textcolor{gray}{66.8} & \textcolor{gray}{27.2} & \textcolor{gray}{48.2} & \textcolor{gray}{32.4} & \textcolor{gray}{31.2} & \textcolor{gray}{38.3} & \textcolor{gray}{41.5 (+22.7)} \\
\midrule
Faster R-CNN w/ Rot  &VGG-16& \textbf{42.2} & \textbf{47.2}  & \textbf{59.8} & 23.2 & \textbf{43.5} & 19.8 & 27.2 & \textbf{40.0} & \textbf{37.8 (+19.0)}\\
\midrule
Faster R-CNN (source only)~\cite{hsu2020every} &\multirow{3}{*}{ResNet-101} & 33.8 & 34.8 & 39.6 & 18.6 & 27.9 & 6.3 & 18.2 & 25.5 & 25.6 \\
EPM~\cite{hsu2020every} && 41.5 & 43.6 & 57.1 & \textbf{29.4} & 44.9 & \textbf{39.7} & 29.0 & 36.1 & 40.2 (+14.6) \\
\textcolor{gray}{Faster R-CNN (oracle)} && \textcolor{gray}{52.3}  & \textcolor{gray}{55.8}& \textcolor{gray}{73.8} & \textcolor{gray}{37.8} & \textcolor{gray}{54.4} & \textcolor{gray}{31.3} & \textcolor{gray}{36.4} & \textcolor{gray}{47.3} & \textcolor{gray}{48.6 (+23.0)}\\
\midrule 
Faster R-CNN w/ Rot &ResNet-101& \textbf{45.8} & \textbf{51.0} & \textbf{63.1} & 26.8 & \textbf{47.1} & 23.6 & \textbf{30.6} & \textbf{43.6} & \textbf{41.5 (+15.9)} \\
\bottomrule
\end{tabular}}}
\vspace{-4mm}
\end{table*}

\subsection{Evaluation Results}
\label{sec:uda_res}
\minisection{Weather adaptation.} 
We evaluate \model w/ rotation on the Cityscapes$\rightarrow$ Foggy Cityscapes benchmark and show the results in Table~\ref{tab:main_city2fog}.
Our model consistently outperforms pervious approaches with both VGG-16 and ResNet-101 as backbones. We can achieve an mAP of
41.5 points, improving the base Faster R-CNN model by 15.9 points. It
also outperforms the prior art, EPM, by 1.3 points. Similar results apply to using VGG-16 as backbone. \model
improves the base Faster R-CNN from 18.8 to 37.8, an absolute improvement of 19 points in mAP. It also outperforms the prior art by $\sim$2 points. These results indicate that \model w/ Rot can effectively
handle the domain shift despite its architectural simplicity.

We also report the average precision of each category. We consistently improve the prior art EPM for all classes except \emph{truck} and \emph{train}, which have limited annotations available (466 and 158 annotations).
More complicated adaptation strategies might be useful for these rare classes and we leave detailed study for future work.

\minisection{Synthetic-to-real.}
We additionally compare our method with previous approaches under the synthetic-to-real setting shown in Table~\ref{tab:sim2real}. The model is trained on the Sim10K dataset with full annotation together with the unlabeled data from the training set of Cityscapes. We evaluate on the validation set of Cityscapes and report the mAP50 for \emph{car}, as this is the only class in Sim10k dataset.

Our model outperforms all the other methods with both VGG-16 and ResNet-101 as backbones. With ResNet-101 as our backbone, we improve the prior art EPM from 51.2 to 52.4. With VGG-16 as backbone, our model can also improve upon EPM by 1.1 points. 

\begin{table}[tp]
\centering
\small{
\setlength{\tabcolsep}{0.9em}
\caption{\small Adapt from Sim10K to Cityscapes (S$\rightarrow$C). \model w/ Rot outperforms the previous approaches consistently. \vspace{-3mm}}
\label{tab:sim2real}
\adjustbox{width=\linewidth}{
\begin{tabular}{lcl}
\toprule
Method& Backbone& S$\rightarrow$C (mAP50) \\
\midrule
Faster R-CNN (source only) & \multirow{8}{*}{VGG-16} & 30.1 \\ 
DAF~\cite{chen2018domain} && 39.0 (+8.9)\\ 
MAF~\cite{he2019multi} && 41.1 (+11.0) \\ 
SW-DA~\cite{saito2019strong} && 42.3 (+12.2)\\ 
SW-DA*~\cite{saito2019strong} && 47.7 (+17.6) \\ 
SC-DA~\cite{zhu2019adapting} && 43.0 (+12.9) \\ 
EPM~\cite{hsu2020every} && 49.0 (+18.9) \\ 
\textcolor{gray}{Faster R-CNN (oracle)} & & \textcolor{gray}{69.7 (+39.6)} \\ 
\midrule
Faster R-CNN w/ Rot  & VGG-16&  \textbf{50.1 (+20.0)}\\ 
\midrule
Faster R-CNN (source only) &\multirow{3}{*}{ResNet-101} & 41.8 \\ 
EPM~\cite{hsu2020every} && 51.2 (+9.6) \\ 
\textcolor{gray}{Faster R-CNN (oracle)} &&  \textcolor{gray}{70.4 (+28.6)} \\ 
\midrule
Faster R-CNN w/ Rot &ResNet-101&  \textbf{52.4 (+10.4)}  \\
\bottomrule
\end{tabular}}}
\vspace{-3mm}
\end{table}

\begin{figure}[tp]
  \centering
  \includegraphics[width=\linewidth]{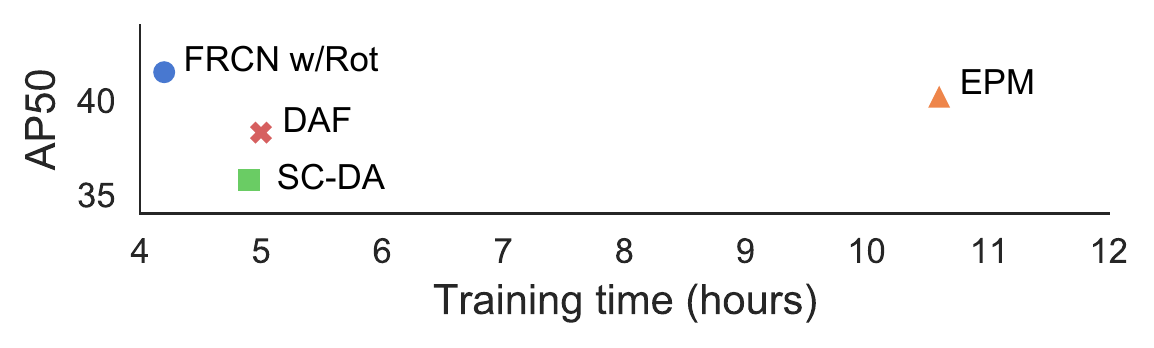}
  \vspace{-6mm}
  \caption{\small Training time and AP of various approaches. We train the previous approaches with ResNet-101 as the backbone on 8 GPUs for weather adaptation. \model w/ Rot reduces the training time of EPM by half while achieving higher AP scores.\vspace{-5mm}}
  \label{fig:runtime}
\end{figure}

\minisection{Training time comparison.} We provide an estimation of the training time of various approaches in Figure~\ref{fig:runtime} by running the
released source code of previous approaches in the same environment. We find that our method reduces the training time of EPM by half while achieving higher AP scores, which is largely due to the simplicity of our approach.

\section{Conclusion}
\vspace{-1mm}
In this work, we investigated the usage of
auxiliary self-supervised tasks to improve the
robustness of object detectors under domain
shifts. Through extensive study on various benchmarks, we show that auxiliary self-supervised
tasks, used in tandem with the object detection training, are effective to improve the out-of-domain
generalization of structural prediction tasks
and that they can be used across different settings, whether abundant test domain data are available. The results can be inspiring 
as a general direction to improve the model robustness under distribution shifts. 
We also introduced instance-level cycle confusion (\task), a self-supervised task on the region features produced by the object detector.
For each object, the task is to find the most different object proposals in the adjacent frame in a video and then cycle back to itself for self-supervision.
\task encourages the object detector to explore invariant structures across instances under various motion, viewpoint, lightning, etc., which leads to improved model robustness in unseen domains at test time.
Our model establishes a new state-of-the-art on large scale video benchmarks.

\vspace{-2mm}
\small{
\subsubsection*{Acknowledgments}
This work was supported by RISE Lab, Berkeley AI Research, Berkeley DeepDrive and DARPA. 
This work was supported in part by DoD including DARPA's XAI, LwLL, and/or SemaFor programs, as well as BAIR's industrial alliance programs. In addition to NSF CISE Expeditions Award CCF-1730628, this research was supported by gifts from Alibaba, Amazon Web Services, Ant Financial, CapitalOne, Ericsson, Facebook, Futurewei, Google, Intel, Microsoft, Nvidia, Scotiabank, Splunk and VMware. Prof. Wang’s group was supported, in part, by gifts from Qualcomm and TuSimple.}

{\small
\bibliographystyle{ieee_fullname}
\bibliography{reference}
}

\appendix
\section*{Appendix}
In this appendix, we provide the detailed results and analysis of the out-of-domain evaluation
benchmarks. In Section~\ref{sec:ood}, we also show the per-category
results of the benchmark results to complement the results in the Table 2 and Table 3
of the main paper and other related evaluations on the Waymo open dataset. In Section~\ref{sec:preds}, we provide more visualizations of the prediction results. 

\section{Out-of-Domain Evaluation Results} 
\label{sec:ood}
In this section, we provide extensive evaluation results of our approach on various datasets and evaluation settings. We find our approach consistently improve over the baseline approaches and the evaluation results on a range of datasets and evaluation settings verify the generality of our method. 

\minisection{Per-category results on BDD100K and Waymo.} In this section, we provide
per-category evaluation results to complement the
Table 2 and 3 in the main paper. We also indicate
the number of instances in each category to give
the readers an idea of the evaluation data
distribution.

In Table~\ref{table:sup_bdd_d2n} and Table~\ref{table:sup_bdd_n2d}, we provide the per-category evaluation results for BDD100K Daytime $\rightarrow$ Night and BDD100K Night $\rightarrow$ Daytime, which is originally shown in Table 2 of the main paper. Our \task task is able to outperform all other tasks across both settings. On BDD100K Daytime $\rightarrow$ Night, \task can achieve at least 1 point improvement in both AP50 and AP75 over other methods. We can observe improvements in several rare categories, including bus, truck, bicycle, and motorcycle. On BDD100K Night $\rightarrow$ Daytime, \task can achieve significant improvements on car and motorcycle categories.

\begin{table*}[h!]
\centering
\small{
\setlength{\tabcolsep}{0.9em}
\caption{\small BDD100K Daytime $\rightarrow$ Night (complement to the result of Table 2 in the main paper).\vspace{-4mm}}
\label{table:sup_bdd_d2n}
\adjustbox{width=\textwidth}{
\begin{tabular}{@{}lccccccccccc@{}} \\\toprule
Model & AP & AP50 & AP75 & person & rider & car & bus & truck & bicycle & motorcycle & train \\\midrule
\# Instances & & & & 12606 & 737 & 107531 & 1760 & 4033 & 846 & 130 & 63 \\
FRCN & 17.84 & 31.34 & 17.68 & 30.62 & 13.19 & 41.39 & 14.36 & 23.38 & 11.46 & 8.38 & 0.00 \\
+Rot & 18.58 & 32.95 & 18.15 & \textbf{30.76} & \textbf{14.39} & 41.38 & 14.17 & 23.07 & 11.62 & 13.24 & 0.00 \\
+Jigsaw & 17.47 & 31.22 & 16.81 & 29.86 & 13.05 & 41.24 & 14.07 & 21.91 & 11.23 & 8.38 & 0.00 \\
+Cycle Consist. & 18.35 & 32.44 & 18.07 & 30.19 & 12.61 & \textbf{42.57} & 15.49 & 22.82 & 11.03 & 12.12 & 0.00 \\
+\task & \textbf{19.09} & \textbf{33.58} & \textbf{19.14} & 30.68 & 13.73 & 41.73 & \textbf{16.71} & \textbf{24.35} & \textbf{12.00} & \textbf{13.53} & 0.00 \\
\bottomrule
\end{tabular}}}
\end{table*}

\begin{table*}[h!]
\centering
\small{
\setlength{\tabcolsep}{0.9em}
\caption{\small BDD100K Night $\rightarrow$ Daytime (complement to the results of Table 2 in the main paper).\vspace{-4mm}}
\label{table:sup_bdd_n2d}
\adjustbox{width=\textwidth}{
\begin{tabular}{@{}lccccccccccc@{}} \\\toprule
Model & AP & AP50 & AP75 & person & rider & car & bus & truck & bicycle & motorcycle & train \\\midrule
\# Instances & & & & 41886 & 1695 & 200372 & 6110 & 21274 & 3047 & 770 & 245 \\
FRCN & 19.14 & 33.04 & 19.16 & 29.63 & 12.90 & 46.55 & 22.12 & 16.82 & 14.03 & 11.04 & 0.00 \\
+Rot & 19.07 & 33.25 & 18.83 & 29.61 & 13.92 & 46.70 & {22.67} & {16.29} & \textbf{14.10} & 9.30 & 0.00 \\
+Jigsaw & 19.22 & 33.87 & 18.72 & {30.03} & 13.68 & 47.01 & 21.68 & 16.49 & 13.94 & 10.97 & 0.00 \\
+Cycle Consist.  & 18.89 & 33.50 & 18.31 & \textbf{30.12} & {13.21} & 47.13 & 22.05 & 17.43 & {13.24} & 7.96 & 0.00 \\
+\task & \textbf{19.57} & \textbf{34.34} & \textbf{19.26} & 29.83 & \textbf{13.95} & \textbf{47.80} & \textbf{23.54} & \textbf{17.10} & 11.58 & \textbf{12.80} & 0.00 \\
\bottomrule
\end{tabular}}}
\end{table*}

 In Table~\ref{table:sup_waymo_fl2n} and Table~\ref{table:sup_waymo_fr2n}, we provide the per-category results for Waymo Front Left $\rightarrow$ BDD100K Night and Waymo Front Right $\rightarrow$ BDD100K Night to complement the results in Table 3 of the main paper. Across both settings, \task outperforms all other methods by a large margin. On Waymo Front Left $\rightarrow$ BDD100K Night, \task can achieve at least 1.5 points improvement in both AP50 and AP75 over other methods, improving the performance on vehicles and pedestrians. On Waymo Front Right $\rightarrow$ BDD100K Night, \task improves the performance on both vehicles and cyclists, while achieving competitive performance with FRCN + Rot on pedestrians.
 
\minisection{Additional cross-camera evaluation on Waymo.} In Table~\ref{table:waymo_ffl} and Table~\ref{table:waymo_fsl}, we provide results for additional settings on the Waymo dataset, Waymo Front $\rightarrow$ Front Left and Waymo Front $\rightarrow$ Side Left. In these settings, the domain gap is due to the change in camera angles. We observe significant improvements for \task on both settings.
On Waymo Front $\rightarrow$ Front Left, \task can achieve around 2 points improvement in AP50 and AP75 and around 4 points improvement in AP for cyclists, while achieving competitive performance on AP for vehicles and pedestrians.
The other self-supervised methods can not obtain improvements.
On Waymo Front $\rightarrow$ Side Left, \task can achieve around 1 point improvement in AP75 and in AP for cyclists.

\begin{table*}[h!]
\centering
\small{
\setlength{\tabcolsep}{0.9em}
\caption{\small Waymo Front Left $\rightarrow$ BDD100K Night (complement to the results in Table 3 of the main paper).\vspace{-4mm}}
\label{table:sup_waymo_fl2n}
\adjustbox{width=.65\linewidth}{
\begin{tabular}{@{}lcccccc@{}} \\\toprule
Model & AP & AP50 & AP75 & vehicle & pedestrian & cyclist \\\midrule
\# Instances & & & & 123749 & 12884 & 737 \\
FRCN & 10.07 & 19.62 & 9.05 & 19.41 & 10.12 & 0.69 \\
+Rot & 11.34 & 23.12 & 9.65 & 21.33 & 12.02 & 0.68 \\
+Jigsaw & 9.86 & 19.93 & 8.40 & 20.40 & 8.78 & 0.41 \\
+Cycle Consist.   & 11.55 & 23.44 & 10.00 & 22.34 & 11.27 & 1.04 \\
+\task & \textbf{12.27} & \textbf{26.01} & \textbf{10.24} & \textbf{22.91} & \textbf{12.70} & \textbf{1.19} \\ \bottomrule
\end{tabular}}}
\end{table*}

\begin{table*}[h!]
\centering
\small{
\setlength{\tabcolsep}{0.9em}
\caption{\small Waymo Front Right $\rightarrow$ BDD100K Night (complement to the results in Table 3 of the main paper).\vspace{-4mm}}
\label{table:sup_waymo_fr2n}
\adjustbox{width=.65\linewidth}{
\begin{tabular}{@{}lcccccc@{}} \\\midrule
Model & AP & AP50 & AP75 & vehicle & pedestrian & cyclist \\\midrule
\# Instances & & & & 123749 & 12884 & 737 \\
FRCN & 8.65 & 17.26 & 7.49 & 17.89 & 7.64 & 0.42 \\
+Rot & 9.25 & 18.48 & 8.08 & 18.22 & \textbf{9.28} & 0.26 \\
+Jigsaw & 8.34 & 16.58 & 7.26 & 16.64 & 8.25 & 0.13 \\
+Cycle Consist.   & 9.11 & 17.92 & 7.98 & 18.87 & 7.80 & 0.65 \\
+\task & \textbf{9.99} & \textbf{20.58} & \textbf{8.30} & \textbf{20.09} & 9.15 & \textbf{0.73} \\ \bottomrule
\end{tabular}}}
\end{table*}

\begin{table*}[h!]
\centering
\small{
\setlength{\tabcolsep}{0.9em}
\caption{\small Waymo Front $\rightarrow$ Front Left.\vspace{-4mm}}
\label{table:waymo_ffl}
\adjustbox{width=.65\linewidth}{
\begin{tabular}{@{}lcccccc@{}} \\\toprule
Model & AP & AP50 & AP75 & vehicle & pedestrian & cyclist \\\midrule
\# Instances & & & & 297909 & 87221 & 1518 \\
FRCN & 36.05 & 57.73 & 38.27 & \textbf{42.08} & 36.99 & 29.08 \\
+Rot & 35.96 & 57.82 & 38.33 & 41.86 & 36.87 & 29.16 \\
+Jigsaw & 35.89 & 57.54 & 38.21 & 41.80 & \textbf{37.05} & 28.81 \\
+Cycle Consist.   & 35.44 & 56.75 & 37.79 & 41.90 & 36.89 & 27.51 \\
+\task & \textbf{37.35} & \textbf{59.78} & \textbf{40.25} & 41.98 & 36.91 & \textbf{33.15} \\ \bottomrule
\end{tabular}}}
\end{table*}

\begin{table*}[h!]
\centering
\small{
\setlength{\tabcolsep}{0.9em}
\caption{\small Waymo Front $\rightarrow$ Side Left.\vspace{-4mm}}
\label{table:waymo_fsl}
\adjustbox{width=.65\linewidth}{
\begin{tabular}{@{}lcccccc@{}} \\\toprule
Model & AP & AP50 & AP75 & vehicle & pedestrian & cyclist \\\midrule
\# Instances & & & & 283889 & 52938 & 1001 \\
FRCN & 31.92 & 53.95 & 33.31 & 36.69 & 29.57 & 29.48 \\
+Rot & 32.66 & 54.29 & 34.33 & \textbf{36.75} & 29.49 & 31.75 \\
+Jigsaw & 32.18 & 53.38 & 33.50 & \textbf{36.75} & \textbf{29.78} & 30.02 \\
+Cycle Consist.   & 31.81 & 53.56 & 33.38 & 36.72 & 29.40 & 29.32 \\
+\task & \textbf{32.89} & \textbf{54.56} & \textbf{35.27} & 36.42 & 29.51 & \textbf{32.74} \\ \bottomrule
\end{tabular}}}
\end{table*}

\section{Visualizations of Prediction Results} 
\label{sec:preds}

\begin{figure*}[h!]
    \centering
    \begin{subfigure}[b]{\textwidth}
        \centering
        \includegraphics[width=\textwidth]{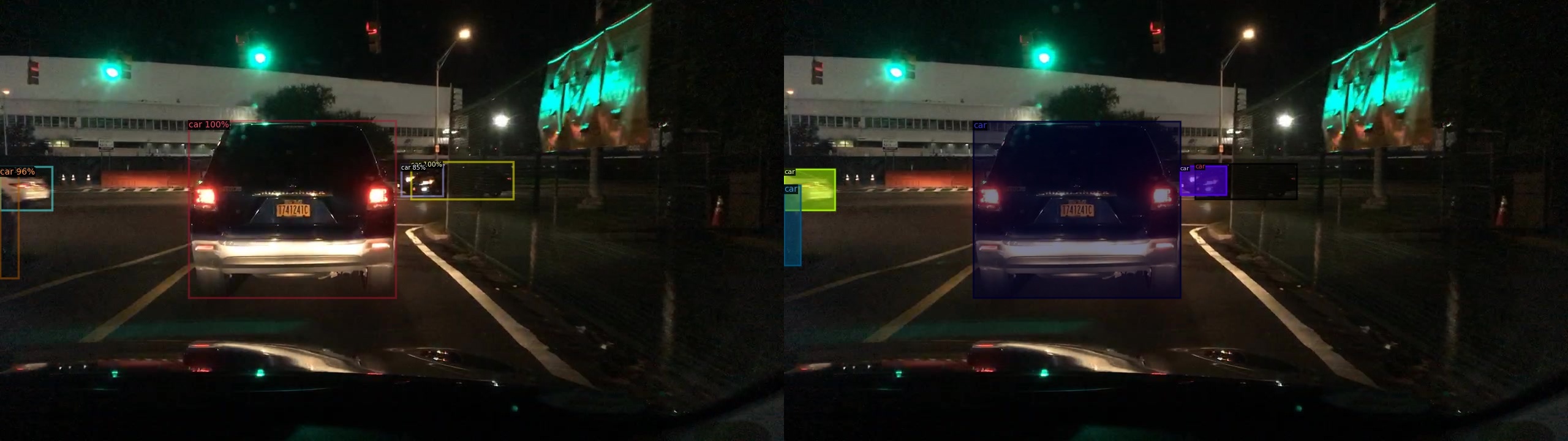}
        \caption{BDD100K Night Seq. 1}
    \end{subfigure}
    \begin{subfigure}[b]{\textwidth}
        \centering
        \includegraphics[width=\textwidth]{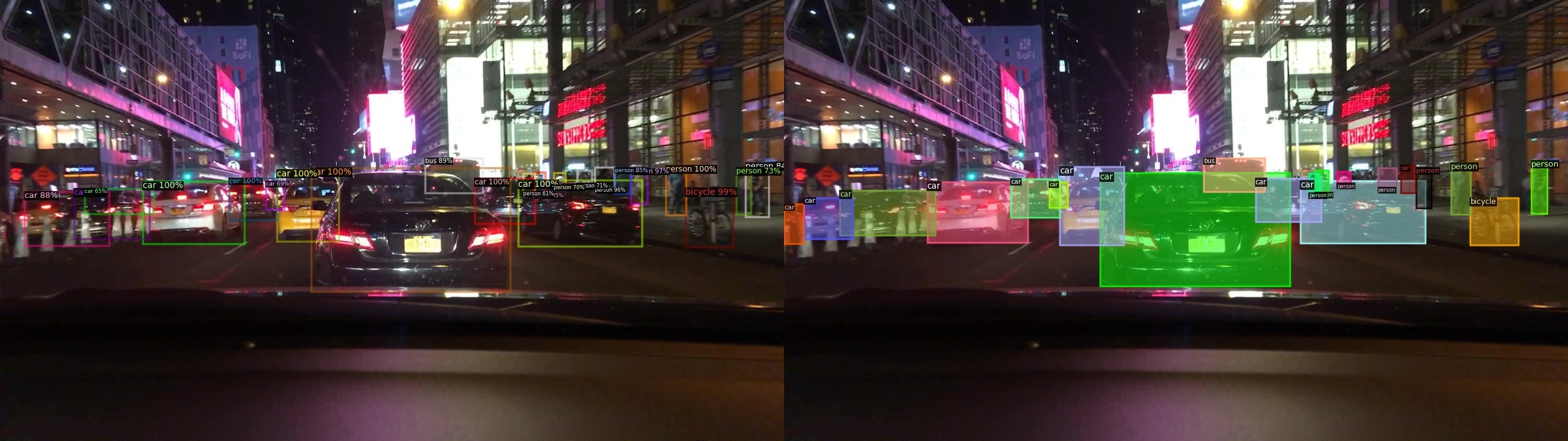}
        \caption{BDD100K Night Seq. 2}
    \end{subfigure}
    \begin{subfigure}[b]{\textwidth}
        \centering
        \includegraphics[width=\textwidth]{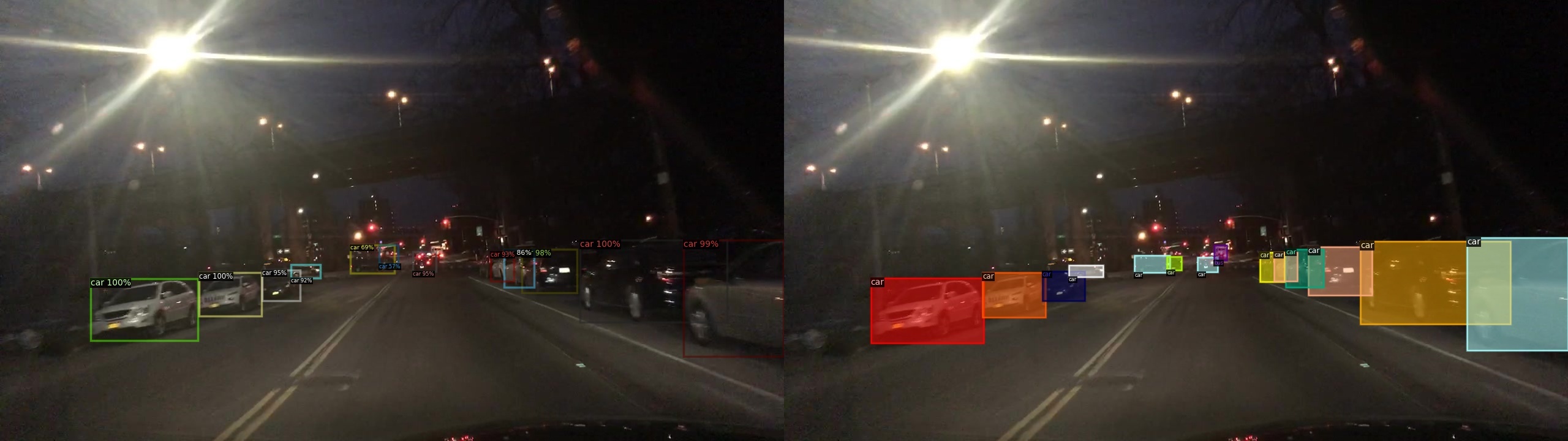}
        \caption{BDD100K Night Seq. 3}
    \end{subfigure}
    \vspace{-2mm}
    \caption{\small Visualization of the predictions of Faster R-CNN w/ \task from BDD100K Daytime to Night. The left side is the model prediction and the right side is the ground truth labels. \vspace{-3mm}}
    \label{fig:visualization}
\end{figure*}

We visualize predictions of \task trained on BDD100K Daytime on frames from several video sequences of BDD100K Night in Figure~\ref{fig:visualization}. Although our model does not observe nighttime images during training, it is still able to successfully identity a majority of the ground truth labels, especially in the densely populated areas.

\begin{figure*}[h!]
    \centering
    \begin{subfigure}[b]{\textwidth}
        \centering
        \includegraphics[width=\textwidth]{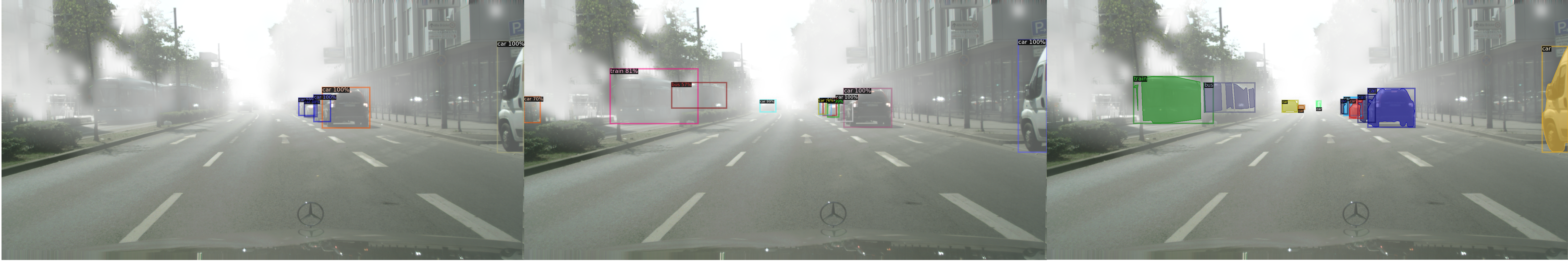}
        
    \end{subfigure}
    \begin{subfigure}[b]{\textwidth}
        \centering
        \includegraphics[width=\textwidth]{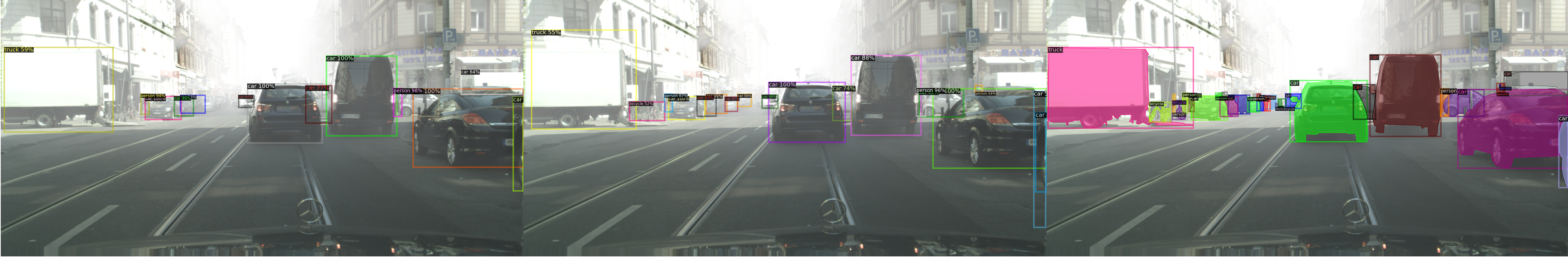}
    \end{subfigure}
    \vspace{-2mm}
    \caption{\small Visualization of UDA experiment on Cityscape dataset. From left to right are the results of baseline, w/rotation and ground-truth. We can observe the model trained with self-supervised task can better find the small or distant objects in the foggy weather, which may be ignored by the baseline methods. \vspace{-3mm}}
    \label{fig:visualization_uda}
\end{figure*}

In Figure~\ref{fig:visualization_uda}, we show the visualization of UDA experiment on Cityscape dataset. From the left to right, we provide the prediction results of the baseline model, the detector trained with rotation and the ground truths. We can observe a more robust prediction of our model under severe distribution shifts. 

\end{document}